\documentclass{article}

 \usepackage[preprint]{neurips_2026}

\usepackage[utf8]{inputenc} %
\usepackage[T1]{fontenc}    %
\usepackage{hyperref}       %
\usepackage{url}            %
\usepackage{booktabs}       %
\usepackage{amsfonts}       %
\usepackage{nicefrac}       %
\usepackage{microtype}      %
\usepackage{xcolor}         %

\usepackage[table]{xcolor} %
\usepackage{siunitx}
\usepackage{multirow}
\usepackage{arydshln}
\usepackage{CJKutf8}

\usepackage{amsthm}
\theoremstyle{plain}
\newtheorem{theorem}{Theorem}[section]

\theoremstyle{definition}
\newtheorem{definition}[theorem]{Definition}

\theoremstyle{remark}

\usepackage{enumitem}
\usepackage[most]{tcolorbox}
\usepackage{tabularray}
\usepackage{array}

\newcolumntype{L}{>{\raggedright\arraybackslash}X}

\newtcolorbox{findingsbox}{
    colback=gray!15,           %
    colframe=gray!15,          %
    arc=3mm,                   %
    boxrule=0pt,               %
    left=10pt,                 %
    right=10pt,                %
    top=10pt,                  %
    bottom=5pt,               %
    enhanced,                  %
    breakable                  %
}

\title{Exploring the AI Obedience: 
Why is Generating \\ a Pure Color Image Harder than CyberPunk?}

\author{%
  \textbf{Hongyu Li}\thanks{Equal contribution.}\, $^{1}$ , \textbf{Kuan Liu}$^{*2}$, \textbf{Yuan Chen}$^{2}$, \textbf{Juntao Hu}$^{2}$, \textbf{Huimin Lu}$^{2}$, \\
  \textbf{Guanjie Chen}$^{3}$, \textbf{Hong Huang}$^{2}$, \textbf{Guangming Lu}$^{1}$, \textbf{Xue Liu}, \textit{IEEE Fellow}$^{4}$ \\
  \vspace{0.05in} \\
  $^1$Harbin Institute of Technology, Shenzhen \quad $^2$City University of Hong Kong \\
  $^3$Chinese University of Hong Kong \quad $^4$McGill University \\[0.2cm]
  \vspace{0.1in}
  \centering
  \small
  \makebox[\textwidth][c]{
    \raisebox{-0.8mm}{\includegraphics[height=13pt]{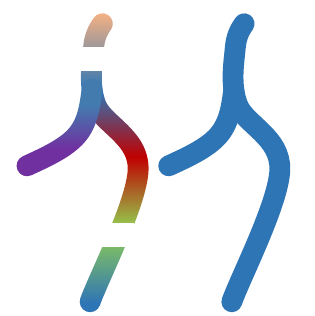}} \hspace{1pt} \href{https://ai-obedience.github.io}{\color[RGB]{0, 51, 153}Project} \quad
    \raisebox{-0.8mm}{\includegraphics[height=13pt]{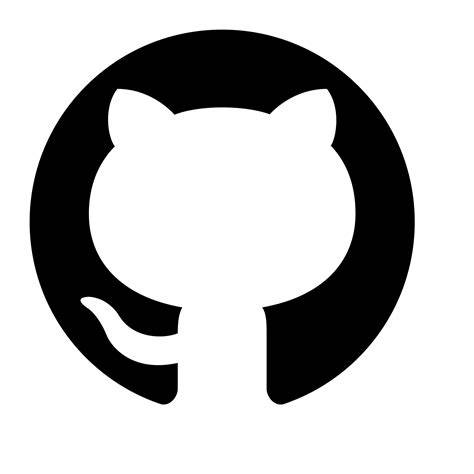}} \hspace{1pt} \href{https://github.com/AI-Obedience/Violin}{\color[RGB]{0, 51, 153}GitHub} \quad
    \raisebox{-0.8mm}{\includegraphics[height=13pt]{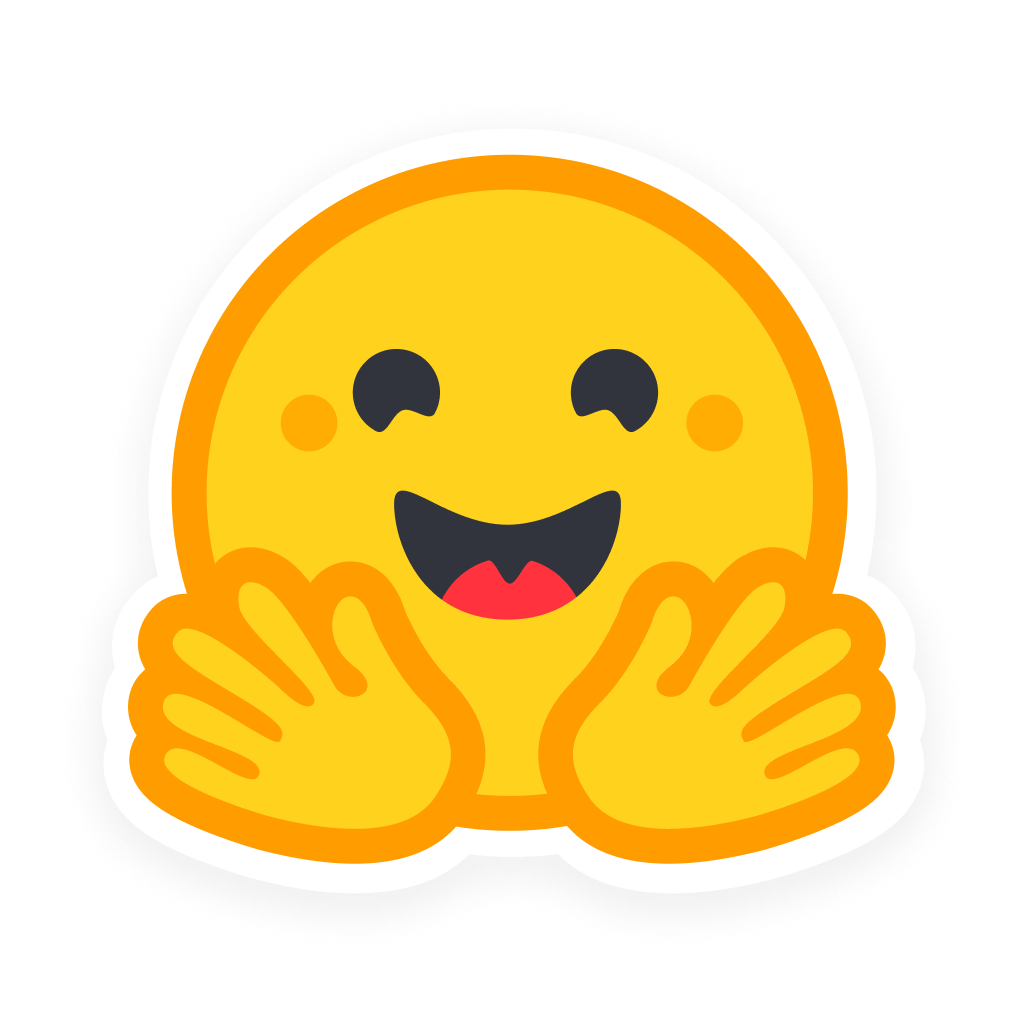}} \hspace{1pt} \href{https://huggingface.co/datasets/Perkzi/VIOLIN}{\color[RGB]{0, 51, 153}HuggingFace}
  }
}

\begin{document}

\maketitle

\begin{abstract}
  Recent advances in generative AI have shown human-level performance in complex content creation. 
  However, we identify a "Paradox of Simplicity": models that can render complex scenes often fail at trivial, low-entropy tasks, such as generating a uniform pure color image. 
  We argue this is a systemic failure related to uncontrollable emergent abilities. 
  As models scale, strong priors for aesthetics and complexity override deterministic simplicity, creating an "aesthetic bias" that hinders the model's transition from data simulation to true intellectual abstraction.
  To better investigate this problem, we formalize the concept of AI Obedience, a hierarchical framework that grades a model's ability to transition from probabilistic approximation to pixel-level determinism (Levels 1 to 5).
  We introduce Violin, the first systematic benchmark designed to evaluate Level 4 Obedience through three deterministic tasks: color purity, image masking, and geometric shape generation. 
  Using Violin, we evaluate several state-of-the-art models and reveal that closed-source models generally outperform open-source ones in deterministic precision. Interestingly, performance on our benchmark correlates with the benchmark in natural image generation. 
  Our work provides a foundational framework and tools for achieving better alignment between human instructions and model outputs.
\end{abstract}

\section{Introduction}

The rapid advancement of generative AI has achieved a significant milestone in producing content at a human-level ~\cite{karras2019style, ho2020denoising, rombach2022high, labs2025flux, wu2025qwen}. 
These methods have achieved breakthroughs in fields such as image editing \cite{brooks2023instructpix2pix}, programming \cite{guo2024deepseek, zhu2024deepseek}, and conversational document analysis \cite{grattafiori2024llama}, boosting daily and professional efficiency.
However, although achievement, the current generative AI creates a ``Paradox of Simplicity'': models that can render complex landscapes often fail at trivial tasks. 
For example, most AI failed to generate a pure color image, always tend to add objects, noise, or textures.
This raises a fundamental question: \textit{Why is generating a pure red image harder for AI than a ``Cyberpunk Cityscape''?} 

\begin{figure*}[t]
    \centering 
    \includegraphics[width=0.82\linewidth]{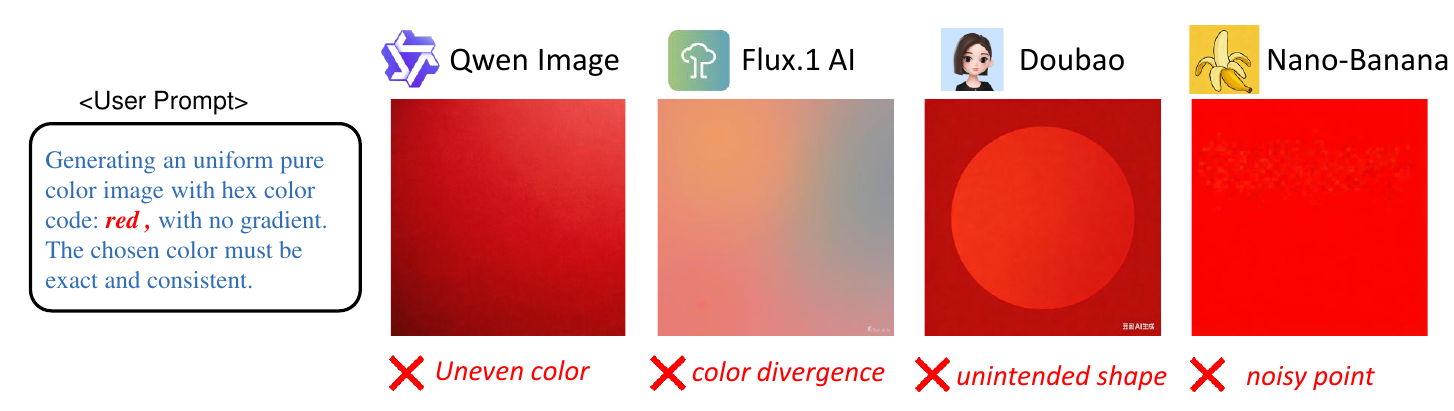}
    \caption{\textbf{Visualization of instruction-following failures} in pure color generation. Instead of adhering strictly to input instructions, models reflexively introduce spurious artifacts and gradients, highlighting a critical bottleneck in precise generative control.}
    \label{fig-moti-tests}
\end{figure*}

We argue that this phenomenon is not a random error, but a critical and systemic failure case. If the model can synthesize hyper-realistic textures, it should possess the capacity for low-entropy patterns. Just as humans can easily imagine "pure colors" rarely found in nature, higher intelligence should possess a similar ability to go beyond data experience and simulation.

Therefore, we attribute the root cause to the uncontrollability of emergent abilities~\cite{wei2022emergent,berti2025emergent,power2022grokking,havlik2025llms}. Under the guidance of scaling laws~\cite{kaplan2020scaling}, large-scale models often exhibit unexpected capabilities and tendencies, such as recidivism prediction~\cite{ganguli2022predictability}. These emergent behaviors suggest that models do not just process data, but internalize complex human orientations; for instance, models tend to develop priors for aesthetics or complexity~\cite{taylor2026algorithmic,doh2026aesthetics,guo2025aesthetic}, favoring intricate content or "pleasing" attributes like symmetry. While occasionally beneficial, these priors often conflict with the deterministic tasks(low-entropy tasks), leading to undesirable results.

To further investigate this issue, we delve deeper into the concept of controllability, which lacks a systematic definition for now. Therefore, we first propose the concept of \textbf{AI Obedience}, a high-level conceptual framework, which defines a level system from conceptual approximation to numerical exactness:
Level 1 (Semantic Obedience) and Level 2 (Relational Obedience) focus on content alignment, ensuring the model captures the correct categories and binds the correct attributes. 
Level 3 (Constraint Obedience) introduces a phase shift: \textit{inhibition}. Here, the model must control its generative priors to follow negative constraints.
Level 4 (Instructional Obedience) and Level 5 (Systemic Obedience) represent the transition to ``pixel-level determinism'', where the model must treat the output not as a probabilistic image, but as a rigid data structure governed by numerical values or coordinates. 
Beyond limiting creativity, the concept of Obedience requires the model to retain flexibility when allowed for open-ended generation.

Obedience is critical for practical applications. 
First, it is essential for safety: if identical prompts yield unpredictable results, AI systems cannot be trusted in automated pipelines. 
Second, in domains like medical imaging, any deviation from instructions introduces data corruption. 
For instance, if a model is asked to ``mark all regions with density above threshold X'' but injects false information into diagnostic data, may causing serious medical accidents.  
While harness engineering has recently explored this area from a similar perspective~\cite{202603.1756}, it focuses primarily on external control mechanisms for agent AI. In contrast, our work aims to investigate the native internal capabilities of the model. By shifting the focus from external constraints to internal mechanisms, we can fundamentally reveal the source of deterministic outputs. Furthermore, this approach provides a potential foundation for the evolution of harness engineering.

Our work focuses on deterministic tasks, specifically L4, with L4 designated as the research foundation for L5. In constructing the benchmark, we carefully balanced the characteristics of natural language processing and computer vision tasks. Given that the fundamental units of vision (pixels) far exceed those of language (tokens) in scale, for instance, a $512 \times 512$ image contains 262,144 pixels, whereas a comparable token count is rare and difficult to evaluate, visual control presents a more rigorous challenge for deterministic generation. Therefore, we select visual tasks as our baseline and propose Violin, which comprises three deterministic obedience tasks: pure color generation, image mask, and geometry shape generation. These tasks maintain zero- or low-entropy characteristics, yielding nearly deterministic outputs. Leveraging the Violin benchmark, we evaluate the obedience capabilities of existing open-source and closed-source models.

Overall, the primary contributions of this work are threefold: 
\begin{itemize}
    \setlength{\itemsep}{0pt}    %
    \setlength{\parsep}{0pt}     %
    \setlength{\parskip}{5pt}    %
    \item \textbf{Obedience Concept and Framework}: We are the first to formalize the concept of AI Obedience, establishing a hierarchical grading system with Level $0\sim5$. Within this framework, we conduct case studies to present key observations and formulate hypotheses.

    \item \textbf{Violin Benchmark:}: To bridge the gap in high-level evaluation, we introduce Violin, the first systematic benchmark dedicated to Level 4 Obedience. This includes a dataset featuring six variations and a dual-metric suite measuring {color precision} and {color purity}.

    \item \textbf{Empirical Analysis and Discovery}: We evaluated existing models using the proposed Violin benchmark, revealing that closed-source models generally outperform open-source models. We also observed a consistent correlation between the metrics on our benchmark and those on natural images, further validating the practical value of Violin.
\end{itemize}

\section{Obedience and Observations}

\subsection{Definition of Obedience}

In the pursuit of \textbf{A}rtificial \textbf{G}enerative \textbf{I}ntelligence (AGI), the focus is shifting from specialized task solvers to versatile, general-purpose systems. However, as these models gain increasing autonomy and generative power, controllability emerges as a fundamental prerequisite for their integration into real-world workflows, where unpredictable outputs can lead to significant functional failures. To emphasize a controllability concept that contrasts with current research trends, which often prioritize creative diversity over strict adherence, we introduce the concept of \textbf{Obedience}. This term is designed to characterize the degree to which model outputs precisely execute input instructions, serving as a rigorous metric for the reliability of AGI systems in following human intent.

\begin{definition}
\textbf{Obedience}: 
Let $\mathcal{M}: \mathcal{I} \rightarrow \mathcal{P}(\mathcal{O})$ be a generative model that maps an instruction space $\mathcal{I}$ to a probability distribution $\mathcal{P}(\cdot)$ over the output space $\mathcal{O}$. For an instruction $I\in \mathcal{I}$ at level $L \in \{1, \dots, 5\}$, the \textbf{Obedience Score} is defined by a level-specific operator, which consists of a suite of specialized metrics designed to verify the various constraints required at that level:

\begin{equation}
\text{Obedience}_{\mathcal{M}}(I, L) = \Phi_L \left( \mathbb{P}_{\mathcal{M}|I}, f^*_L(I) \right)
\end{equation}

where $\mathbb{P}_{\mathcal{M}|I}$ denotes the output distribution generated by the model given instruction $I$, $f^*_L(I)$ defines the oracle mapping for instruction $I$, representing the deterministic ground-truth output that perfectly satisfies the constraints of level $L$. This formulation unifies different evaluation granularities:

\begin{itemize}[parsep=0pt, topsep=1pt]
    \item \textbf{Per-instance level}: $\Phi_L$ acts as a metric $d(o, o^*)$ between a single sample $o \sim \mathbb{P}_{\mathcal{M}|I}$ and the target $f^*_L(I)$.
    \item \textbf{Distribution level}: $\Phi_L$ measures the divergence between the generated distribution $\mathbb{P}_{\mathcal{M}|I}$ and a target distribution to account for generative stochasticity.
\end{itemize}    
\end{definition}

\subsection{Obedience Levels}
\begin{figure*}[t]
    \centering
    \includegraphics[width=0.95\linewidth]{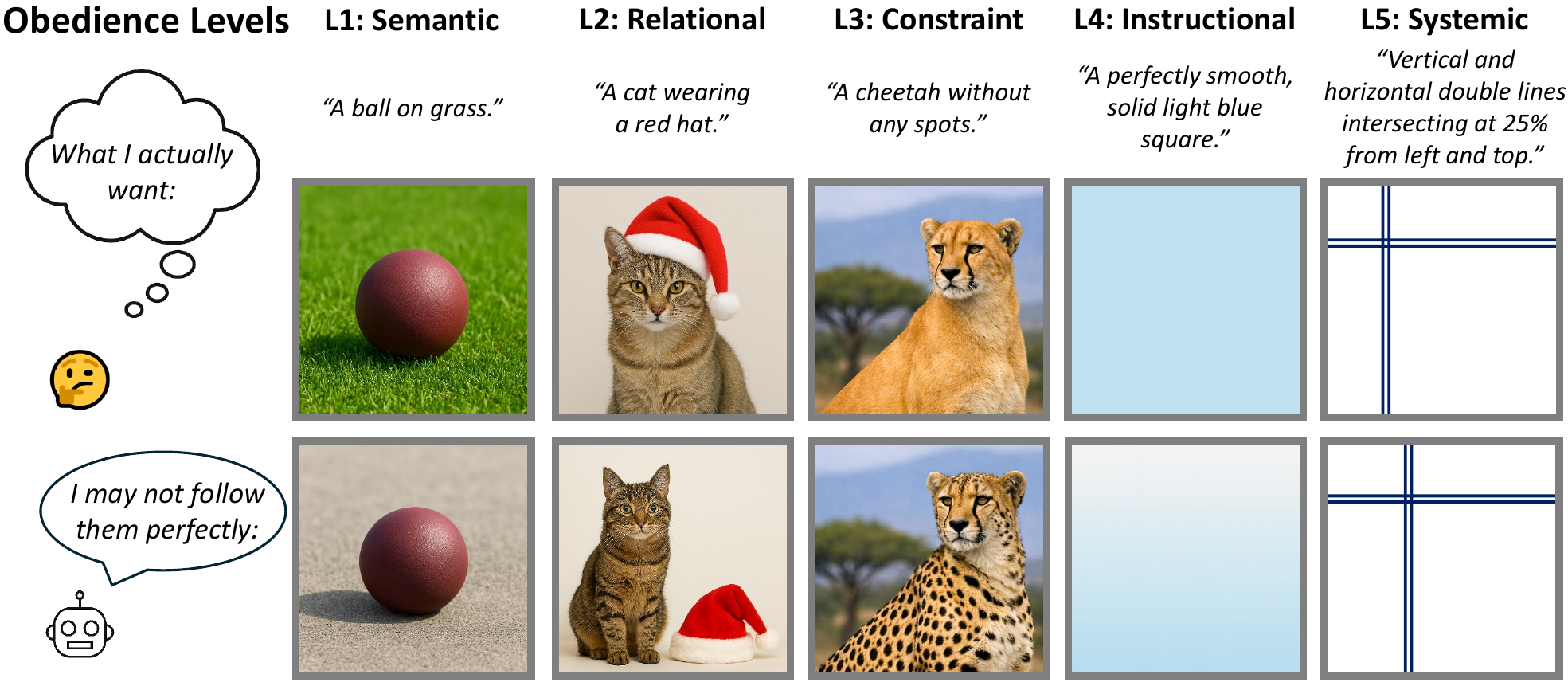}
    \caption{Illustration of obedience levels in the vision task. Each column shows the prompt, the expected output, and a failure case.}
    \label{fig:obedience_levels}
\end{figure*}

To provide a rigorous framework for evaluating generative systems, we propose a grading system of the AI Obedience standard. This hierarchy characterizes the transition from probabilistic approximation to deterministic execution, centered on the model's ability to govern its generative priors. Intuitive examples can be seen in Fig.~\ref{fig:obedience_levels}.

\begin{enumerate}[label=\textbf{Level \arabic*: }, wide=0pt, leftmargin=0pt, labelsep=0.3em, itemsep=0.6em, start=0]
    \item \textbf{Non-Obedience} \\
    The model ignores the user's intent. Whether the instruction is general or specific, the model follows only its internal training data and fails to acknowledge the prompt.
    \begin{itemize}[label=$\diamond$]
        \item \textit{Criterion:} When the instruction is general or ambiguous, the model ignores the prompt's constraints and produces unrelated output.
    \end{itemize}

    \item \textbf{Semantic Obedience (``Vibe'' Level)} \\
    At this level, obedience is defined by the alignment between the instruction's broad intent and the output semantics. 
    \begin{itemize}[label=$\diamond$]
        \item \textit{Criterion:} For general or low-constraint instructions, the model must map broad conceptual requirements to a valid representation within the output domain.
        \item \textit{Failure Mode:} The output falls outside the intended conceptual boundaries, resulting in a mismatch with the prompt's intent (e.g., unrelated objects, incorrect task types, or irrelevant domains)
    \end{itemize}

    \item \textbf{Relational Obedience (``Binding'' Level)} \\
    At this level, obedience is defined by maintaining the integrity of multi-entity instructions, ensuring that specific attributes and relational constraints are correctly localized.
    \begin{itemize}[label=$\diamond$]
        \item \textit{Criterion:} For complex or high-constraint prompts, the model must accurately bind independent properties to their respective subjects and respect relative positioning or logical associations.
        \item \textit{Failure Mode:} Properties or relations migrate between entities, resulting in a structural mismatch (e.g., mismatched attributes, swapped roles, or inverted spatial/logical relationships).
    \end{itemize}

    \item \textbf{Constraint Obedience (``Inhibition'' Level)} \\
    At this level, obedience is defined by the model's capacity to follow restrictive or negative constraints by intentionally suppressing its internal priors.
    \begin{itemize}[label=$\diamond$]
        \item \textit{Criterion:} For exclusionary or minimalist prompts, the model must filter out prohibited elements and maintain a strict adherence to the specified limitations.
        \item \textit{Failure Mode:} The model introduces unrequested attributes (e.g., adding decorative textures, stylistic nuances, or extra logic) to avoid a simplistic output.
    \end{itemize}

    \item \textbf{Instructional Obedience (Deterministic Level)} \\
    At this level, obedience is defined by the elimination of stochastic variance within the output's fundamental units. The model must treat the output space as a rigid data structure, ensuring that every discrete element aligns perfectly with the specified numerical parameters.
    \begin{itemize}[label=$\diamond$, leftmargin=1.5em]
        \item \textit{Criterion:} For accurate (or parameterized) instruction, the model must achieve zero-entropy mapping (e.g., exact color values or bit-perfect consistency) where the output signal is a 1:1 reflection of the instruction.
        \item \textit{Failure Mode:} Subtle fluctuations (e.g., pixel-level noise or value variance) that prevent the output from reaching a pure, mathematically certain state.
    \end{itemize}

    \item \textbf{Systemic Obedience (``Architectural'' Level)}\\
    The highest level of obedience represents the total alignment of generative capacity with complex geometric or logical specifications. The model operates as a high-fidelity rendering engine, strictly adhering to absolute coordinates and rigid structural interdependencies.
    \begin{itemize}[label=$\diamond$]
        \item \textit{Criterion:} For coordinate-based or blueprint-style prompts, the model must ensure geometric exactness and dimensional accuracy, verifiable through external measurement tools.
        \item \textit{Failure Mode:} Misalignment of coordinates or structural warping (e.g., an entity appearing at coordinates $(120, 120)$ instead of the requested $(100, 100)$).
    \end{itemize}
\end{enumerate}

This hierarchy reveals a fundamental shift in how we evaluate generative AI by drawing a clear line between. We define this critical boundary as the ``Exact Alignment Threshold'': \textit{Beyond Level 3, the model transitions from probabilistic interpretation to deterministic execution, achieving the ability to map input instructions to output details with zero-variance precision.}

\textbf{Relation to Existing Concepts:} Obedience serves as a unified framework that categorizes existing evaluation standards based on the strictness of the instructions. We provide the category of current benchmarks in the Appendix:

\begin{figure*}[h]
    \centering
    \includegraphics[width=0.88\linewidth]{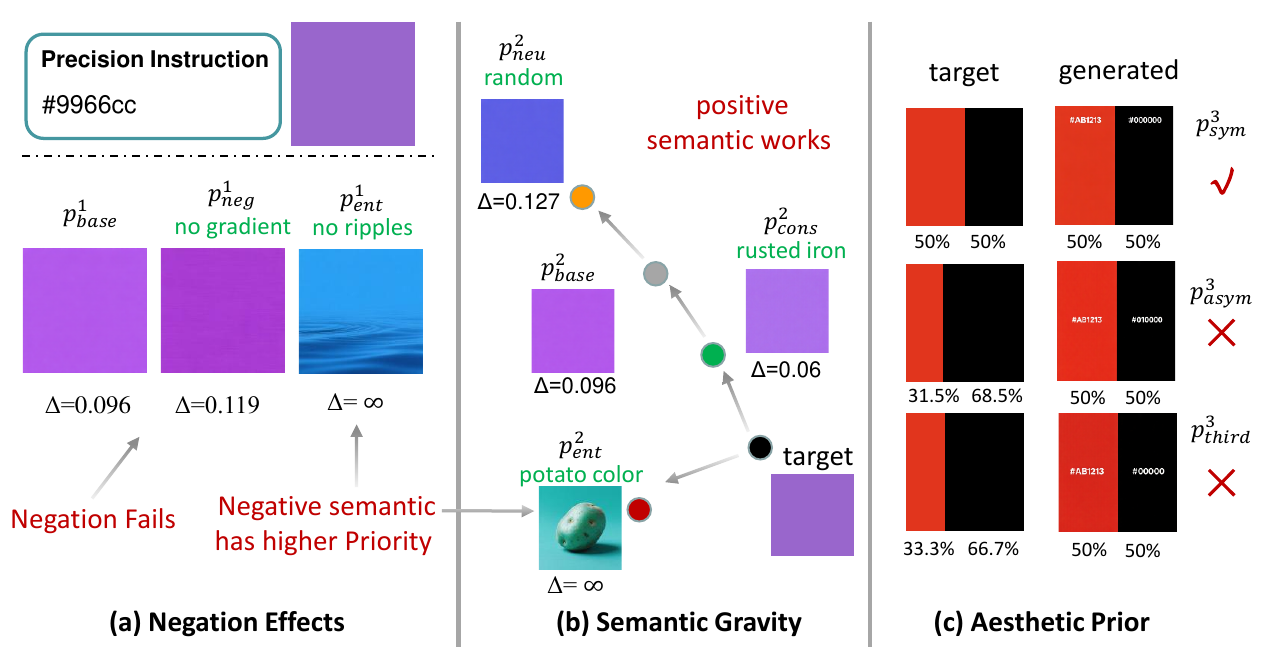}
    \caption{\textbf{Diagnostic case studies}  (a) Logical Inhibition Failure: Negative prompts (``no gradient'') fail to remove artifacts, and mentioning a semantic object to avoid (``ripples'') causes the model to generate instead. (b) Semantic Gravity: The model follows color instructions better when they align with common knowledge (``rusted iron''), but drifts when the context is conflicting or random. (c) Aesthetic Inertia: Precise spatial ratios (31.5\%) are ignored in favor of a standard 50/50 split.}
    \label{fig-case-study}
\end{figure*}

\subsection{Observation}

To identify the bottlenecks in achieving high-level obedience, we identify three aspects: Negation, Semantic Gravity, and Aesthetic Inertia in the pure color generation task using Qwen-Image.

\subsubsection{The effect of Negation}
This study examines the model's capacity for Logical Inhibition. Specifically, we investigate whether the model can utilize negative prompts to suppress unintended color interference.

\begin{findingsbox}
\begin{itemize}[leftmargin=1em, nosep, itemsep=4pt]
    \item $P^1_{\text{base}}$: A simple, positive instruction providing only the hex code (\textit{``Generating a uniform pure color image with hex color code: \#9966CC.''})

    \item $P^1_{\text{neg}}$: Introducing strict negative constraints targeting high-probability textures for gray (\textit{``..., strictly no shadows, no gradients, and no metallic textures.''})

    \item $P^1_{\text{ent}}$: Using terms that are semantically distant but visually noisy (\textit{``..., strictly no cloud patterns or water ripples.''}) 

\end{itemize}
\end{findingsbox}

\subsubsection{The ``Semantic Gravity''}
This study investigates when numerical instructions are subjected to varying semantic contexts, a phenomenon we term ``Semantic Gravity.'' We examine whether the model maintains numerical fidelity or drifts toward the semantic center of the surrounding text.

\begin{findingsbox}
\begin{itemize}[leftmargin=1em, nosep, itemsep=4pt]
    \item $P^2_{\text{base}}$: the same as $P_{\text{1,base}}$.

    \item $P^2_{\text{cons}}$: associate with a consistent prior, testing if aligned priors can act as an anchor to improve precision. (\textit{``..., which is the typical color of a rusted iron plate''}), 

    \item $P^2_{\text{conf}}$: associate with a conflicting prior to measure the ``pull'' of training data distributions. (\textit{``..., representing the color of a fresh potato''})

    \item $P^2_{\text{neu}}$: Explicitly stripping semantic meaning to observe if logical de-biasing can mitigate aesthetic interference. (\textit{``..., a randomly generated color with no specific meaning or real-world counterpart''}) 

\end{itemize}
\end{findingsbox}

\subsubsection{Spatial Precision vs. Aesthetic Bias}
This study investigates the conflict between precise spatial layout and ``Aesthetic Inertia'', the tendency to favor balanced, conventional compositions over exact numerical ratios. We test whether the model can execute spatial instructions when they deviate from common visual patterns:

\begin{findingsbox}
\begin{itemize}[leftmargin=1em, nosep, itemsep=4pt]

    \item $P^3_{\text{sym}}$: A standard balanced split (\textit{``A split image with two solid color blocks: the left 50\% is a pure solid color with hex code \#AB1213, the right 50\% is a pure solid color with hex code \#000000''}).

    \item $P^3_{\text{asym}}$: A non-standard, precise ratio (\textit{``A split image with two solid color blocks: the left 31.5\% is a pure solid color with hex code \#AB1213, the right 68.5\% is a pure solid color with hex code \#000000''}).

    \item $P^3_{\text{third}}$: A common photographic composition (\textit{``A common photographic composition with two solid color blocks: the left 33.3\% is a pure solid color with hex code \#AB1213, the right 66.7\% is a pure solid color with hex code \#000000''}).

\end{itemize}
\end{findingsbox}

\subsubsection{Summary of Observations and Hypotheses.}
As shown in Fig.~\ref{fig-case-study}, our case studies highlight a fundamental gap between logical instructions and visual execution. In case (a), negative constraints backfire, with the model generating specific features (like ripples) that it was explicitly told to avoid. Case (b) demonstrates that while familiar semantic anchors (like "rusted iron") improve precision, irrelevant or confusing prompts lead to visual "hallucinations." Finally, case (c) reveals that precise numerical ratios are often ignored in favor of default symmetrical layouts.

These findings suggest that state-of-the-art models operate as "intuitive artists" rather than precision tools, governed by three key behaviors:
(1) Semantic Priming: Positive concepts override negative "no/avoid" commands.
(2) Semantic Gravity: Precise values drift toward familiar training clusters.
(3) Aesthetic Inertia: Pre-existing layout biases override specific spatial instructions.

\section{Violin Benchmark}

\begin{figure*}[t]
    \centering
    \includegraphics[width=1\linewidth]{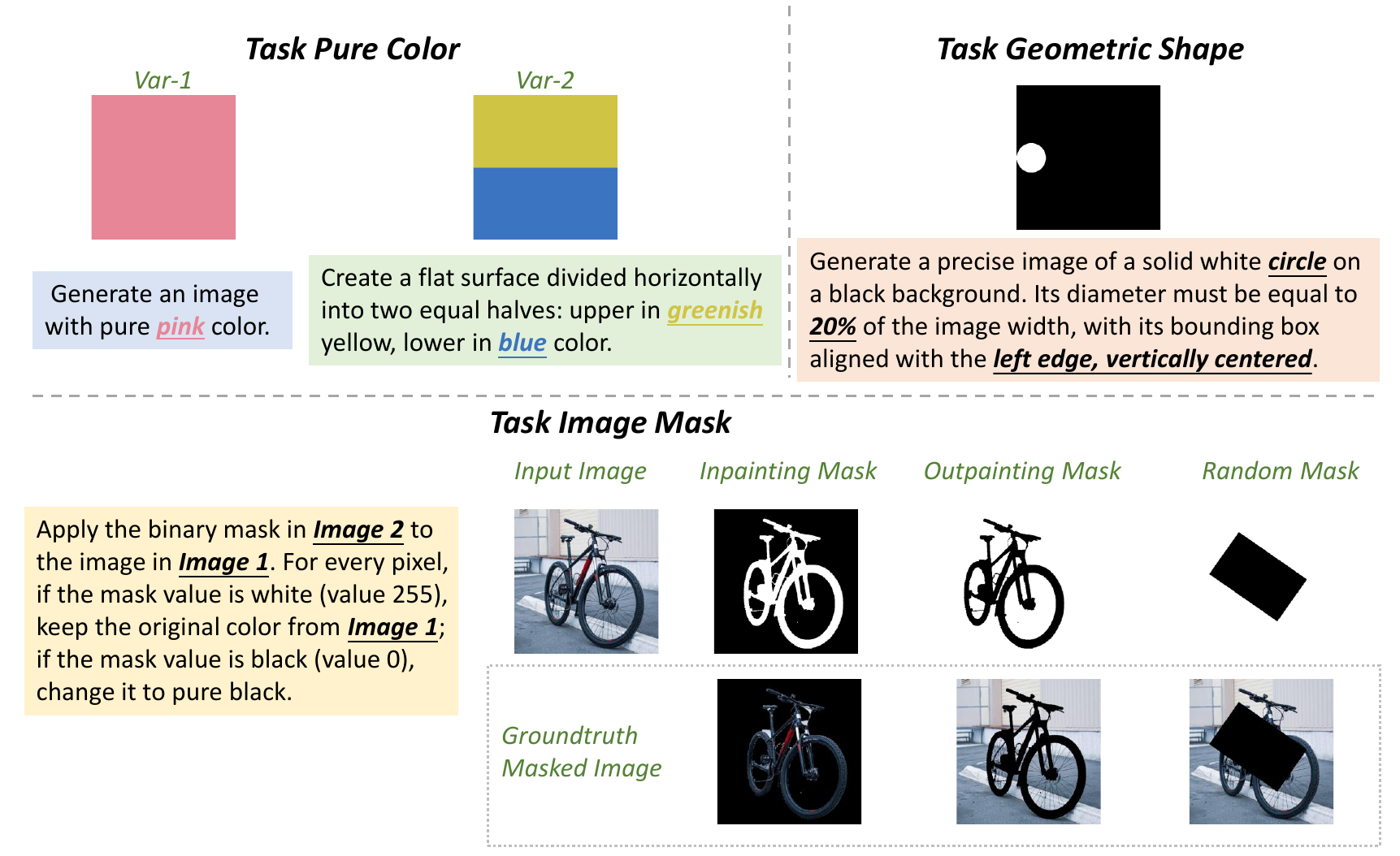}
    \caption{\textbf{Overview of the Violin Benchmark. }The benchmark evaluates Level-4 visual obedience through three tasks: (a) Pure Color Generation; (b) Geometric Shape Generation; and (c) Image Masking for strict spatial accuracy across inpainting, outpainting, and random masks.}
    \label{fig-benchmark-display}
    \vspace{1em}
\end{figure*}

\subsection{Datasets}
To bridge the gap in existing research regarding high-level obedience, we propose the first benchmark specifically designed to assess Level-4 visual obedience: \textbf{Violin} (\textbf{Vi}sual \textbf{o}bedience \textbf{L}evel-4 Evaluat\textbf{i}o\textbf{n}). Expanding beyond mere conceptual generation, Violin serves as a rigorous evaluation tool focused on fundamental visual elements, including pure colors, simple geometries, and image masking. The dataset consists of text-image pairs where the ground truth is programmatically synthesized. Every pixel maintains the strictly requested values, establishing a deterministic standard that eliminates visual ambiguity. The benchmark is organized into three tasks as shown in Fig.~\ref{fig-benchmark-display}.

\textbf{Color Purity Variation 1: Single Color Block.} 
In this basic scenario, the entire image consists of a single uniform color. We utilize natural language color names from the ISCC-NBS System (Level 2) to evaluate the model's ability to execute color instructions across various textual expressions. By excluding all textures, this variation measures the most fundamental deterministic ability.

\textbf{Color Purity Variation 2: Double-Color Block.}
As an extension of Color Variation 1, we construct double-color block tasks to evaluate the model's ability to handle scenarios with spatial contrast. This variation corresponds to two-color blocks with unambiguous spatial partitioning (horizontal or vertical split), utilizing ISCC-NBS Level-2 color names to instruct each region independently.

\textbf{Geometric Shape Generation.}
Moving beyond pure color, this task introduces basic geometric and spatial constraints to evaluate the model's precision in location-based generation. Models are required to generate specific shapes (e.g., circles) at strictly defined positions. This task emphasizes the {spatial alignment} between the prompted position and the generated pixels, assessing whether the model can faithfully execute localized geometric instructions without spatial drifting.

\textbf{Image Masking.}
To evaluate high-level obedience in complex spatial contexts, we introduce the image masking task, which assesses a model's ability to precisely suppress visual content based on an explicit spatial mask. Given an original image, a mask, and a prompt, the model is required to generate an image whose mask strictly covers the target area. We employ three mask configurations: (i) {Inpainting masks}, which target internal objects or semantic regions; (ii) {Outpainting masks}, which target peripheral regions; and (iii) {Random masks}, which test arbitrary spatial adherence.

\subsection{Evaluation Metrics}
To rigorously assess visual obedience, we design specific metrics for each task, each comprising five distinct dimensions. All individual metrics are normalized to $[0,1]$, where higher scores denote increasing deviation from the prompt, indicating worse obedience.

\paragraph{Pure Color Metrics (Var 1 \& 2)}
To quantify the degree of Obedience, we evaluate model performance through two dimensions: {color precision} and {color purity}. For Variation 2, we evaluate each sub-region independently and report the average score. Color Precision quantifies the deviation between generated and target colors, and color purity captures unintended visual artifacts, such as irregular textures and inconsistent gradients.
For color precision: 
(1) Euclidean Distance ($d_{rgb\_ed}$): computes the Euclidean distance across three RGB channels, providing the most intuitive measure.
(2) CIEDE2000 ($d_{lab\_00}$): employs complex nonlinear transformations and incorporates weighting functions to more accurately model the human visual system.
For color purity:
(3) Standard Deviation ($d_{sd}$): compute the standard deviation across all channels.
(4) Canny Edge Density ($ d_{ced}$): employ the canny algorithm to detect edge features and calculate the proportion of edge pixels relative to the total number of pixels.
(5) High-Frequency Component Ratio ($d_{hf}$): transform the image to the frequency domain via 2D Fast Fourier Transform (2D FFT) and compute the energy ratio of high-frequency components.

\paragraph{Geometric Shape Metrics}
To evaluate the ability of synthesizing specific shapes at precise locations with geometric fidelity, we define five metrics:
(1) Shape IoU ($d_{iou}$): measures the global spatial overlap between the generated shape and the ground-truth;
(2) Centroid Offset ($d_{dist}$): quantifies the Euclidean distance between the centroids, normalized by the image diagonal to assess localization precision;
(3) Size Deviation ($d_{size}$): evaluates the relative area error of the generated object compared to the instruction;
(4) Geometric Similarity ($d_{shape}$): utilizes {Hu Moments} (specifically the $I_1$ invariant) to determine whether the topology of the generated object matches the requested shape, ensuring invariance to scale and rotation;
(5) Regional Purity ($d_{pure}$): measures the average color standard deviation strictly within the object's mask to detect internal noise or inconsistent filling.

\paragraph{Image Masking Metrics}
For the image masking task, we assess the model’s performance in accurately applying a pixel-level mask to an existing image:
(1) IoU ($d_{iou}$): computes the global spatial overlap between the generated image and the ground-truth masked image, providing a holistic measure of coverage;
(2) Centroid Offset ($d_{dist}$): quantifies the localization error between the centroids;
(3) Boundary Alignment ($d_{biou}$): specifically evaluates edge precision to penalize boundary blurring or subtle misalignments;
(4) Content Leakage ($d_{leak}$): detects residual information by measuring the average grayscale intensity within the eroded core of the ground-truth mask, ensuring the target is fully obscured;
(5) Edge Sharpness ($d_{edge}$): compares Sobel gradient magnitudes to ensure a crisp, definitive occlusion boundary instead of a soft or feathered transition.

\begin{table*}[t]
    \centering
    \small
    \setlength{\tabcolsep}{10pt}
    \caption{Comparison of Level-4 Obedience on Pure Color Generation Task.}
    \begin{tabular}{llccccc >{\columncolor{gray!15}}c}
        \toprule
        Var & Models & {rgb-ed} & {lab-00} & {sd} & {ced} & {hf} & color-mean \\
        \midrule
        
        \multirow{7}{*}{Var-1} 
        & FLUX.1        & 0.206 & 0.167 & 0.064 & 0.006 & 0.016 & 0.091 \\
        & FLUX.2        & 0.123 & 0.091 & 0.044 & 0.007 & 0.021 & 0.057 \\
        & Z-Image       & 0.135 & 0.092 & 0.078 & 0.002 & 0.007 & 0.061 \\
        & Qwen-Image    & 0.122 & 0.084 & 0.047 & 0.002 & 0.017 & 0.057 \\
        \cdashline{2-8}
        & Nano-Banana-2 & 0.126 & 0.093 & 0.033 & 0.001 & 0.010 & 0.053 \\
        & Seedream-5    & 0.134 & 0.093 & 0.015 & 0.001 & 0.001 & 0.049 \\
        & GPT-Image-2   & 0.137 & 0.083 & 0.006 & 0.000 & 0.031 & 0.051 \\
        \bottomrule
    \end{tabular}
    \label{tab-test-color-var1}
\end{table*}

\begin{table*}[t]
    \centering
    \small
    \setlength{\tabcolsep}{10pt}
    \caption{Comparison of Level-4 Obedience on Image Mask Task.}
    \begin{tabular}{lccccc >{\columncolor{gray!15}}c}
        \toprule
        Models & {iou} & {biou} & {leak} & {edge} & {dist} & shape-mean \\
        \midrule
        FLUX.2         & 0.848 & 0.948  & 0.438 & 0.131 & 0.197 & 0.512 \\
        \cdashline{1-7}
        Nano-Banana-2  & 0.474 & 0.772  & 0.286 & 0.172 & 0.127 & 0.366 \\
        Seedream-5     & 0.331 & 0.725  & 0.177 & 0.073 & 0.093 & 0.280 \\
        GPT-Image-2    & 0.096 & 0.398  & 0.087 & 0.186 & 0.023 & 0.158 \\

        \bottomrule
    \end{tabular}
    \label{tab-test-mask}
\end{table*}

\section{Experiments}

\subsection{Implementation}
\paragraph{Models} To conduct a comprehensive evaluation, we choose recent state-of-the-art visual generation models, covering both open-source and closed-source, as well as T2I architectures and unified multimodal designs. We include several high-performing models, including  FLUX.1~\cite{labs2025flux}, FLux.2~\cite{flux-2-2025}, Z-Image~\cite{cai2025z} and Qwen-Image\cite{wu2025qwen}; For closed-source models, we evaluate industry-leading proprietary systems including GPT-Image-2, Seedream-5, and Nano-Banana-Pro2. 
Additionally, to better assess color purity , we include SANA~\cite{xie2025sana}, Janus-Pro-1.5~\cite{chen2025janus}, OmniGen2~\cite{wu2025omnigen2}, all of which have demonstrated exceptional performance in natural scenes.

\subsection{Visual Level-4 Obedience Evaluation}
To investigate the Level-4 Obedience of existing models, we evaluate three tasks on the full benchmark. Generally, the results demonstrate that closed-source models generally exhibit superior obedience ability compared to open-source models. As illustrated in Tab.~\ref{tab-test-color-var1}, Tab.~\ref{tab-test-mask}, Tab.~\ref{tab-test-geometric}, closed-source models generally outperform their open-source counterparts across all three tasks. Notably, the instruction-following capabilities of the recently released GPT-Image have reached a level nearly indicative of practical utility. Among the open-source models, Flux.2 achieves the best performance, also providing evidence for consistency. Despite being trained on natural image datasets, Flux.2 demonstrates great adaptability on Violin, yielding surprisingly results.

\begin{table*}[t]
    \centering
    \small
    \setlength{\tabcolsep}{12pt} %
    \caption{Comparison of Level-4 Obedience on Geometric Shape Generation Task.}
    \begin{tabular}{lccccc >{\columncolor{gray!15}}c}
        \toprule
        Models & {iou} & {size} & {shape} & {purity} & {dist} & mask-mean \\
        \midrule

        FLUX.1         & 0.725 & 0.703 & 0.167 & 0.172 & 0.267 & 0.407 \\
        FLUX.2         & 0.607 & 0.659 & 0.004 & 0.127 & 0.122 & 0.304 \\
        Z-Image        & 0.648 & 0.686 & 0.002 & 0.133 & 0.409 & 0.376 \\
        Qwen-Image     & 0.645 & 0.630 & 0.001 & 0.283 & 0.128 & 0.337 \\
        \cdashline{1-7}
        Nano-Banana-2  & 0.566 & 0.597 & 0.046 & 0.105 & 0.097 & 0.282 \\
        Seedream-5     & 0.551 & 0.376 & 0.006 & 0.035 & 0.070 & 0.207 \\
        GPT-Image-2    & 0.317 & 0.278 & 0.003 & 0.038 & 0.033 & 0.134 \\

        \bottomrule
    \end{tabular}
    \label{tab-test-geometric}
\end{table*}

\subsection{Consistency between Violin and Other Benchmarks}
\begin{figure*}[t]
    \centering
    \includegraphics[width=0.92\linewidth]{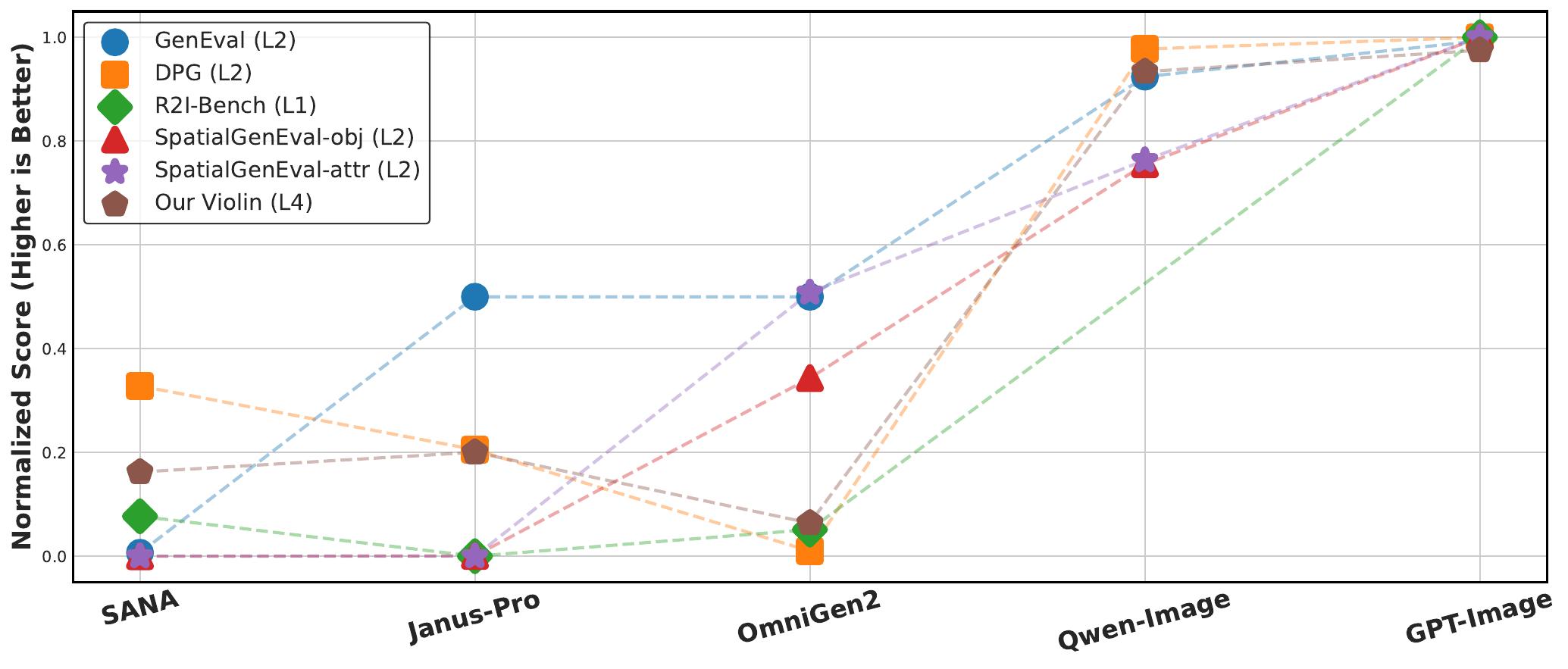}
    \caption{\textbf{Consistency between Violin and Natural Image Benchmark} The proposed VIiolin (L4) maintains a consistent performance trend with existing L1/L2 benchmarks across diverse models.}
    \label{fig-consistency}
    \vspace{-1em}
\end{figure*}

In addition to measuring deterministic output capabilities, our benchmark demonstrates relative consistency. As shown in Fig.~\ref{fig-consistency}, we evaluate our proposed VIOLIN (L4) alongside metrics reported in SpatialGenEval~\cite{wang2026everything}, R2I-Bench~\cite{chen2025r2i}, DPG~\cite{hu2024ella}, and GenEval~\cite{ghosh2023geneval}. The metric values across our three tasks align with those observed in other natural image benchmarks. As such, it serves as a useful, albeit coarse, indicator: strong performance on our benchmark generally suggests similar efficacy on conventional ones.

\section{Conclusions}
In this paper, we have explored the ``Paradox of Simplicity'' in generative AI and formalized the concept of Obedience, establishing a hierarchical grading system. 
Our proposed Violin benchmark, focused on Level-4 Obedience through pure color generation task, image masking task, and geometric shape generation task, offers an effective and robust evaluation and explorative tool. 
Our investigation through case studies and extensive experiments reveals that current generative models often consistently prioritize inherent generative priors or aesthetic priors over logical constraints.
By establishing this framework, we hope to draw more attention to AI Obedience and encourage the development of next-generation models that are not only ``creative artists'' but also ``precise executors''.

\newpage
\bibliographystyle{plainnat}
\bibliography{ref}

\newpage
\appendix

\begin{center}
    {\LARGE \bf Appendix}
\end{center}

\vspace{1cm}

\section*{Contents}
\begin{itemize}
    \item \textbf{Section \ref{sec:relation}: Related Work} \dotfill \pageref{sec:relation}
    \item \textbf{Section \ref{sec:category}: Obedience Levels of Existing Literatures} \dotfill \pageref{sec:category}
    \item \textbf{Section \ref{sec:other-tasks}: Other Obedience Example Tasks} \dotfill \pageref{sec:other-tasks}
    \item \textbf{Section \ref{sec:benchmark}: Benchmark Details.} \dotfill \pageref{sec:benchmark}
    \begin{itemize}
        \item \ref{sec:dataset}: Datasets \dotfill \pageref{sec:dataset}
        \item \ref{sec:metric}: Metrics \dotfill \pageref{sec:metric}
    \end{itemize}
    \item \textbf{Section \ref{sec:ze-color}: Supplementary Zero-Entropy Color Task} \dotfill \pageref{sec:ze-color}
    \begin{itemize}
        \item \ref{sec:violin-absolute-color}: Violin-Absolute-Color Benchmark \dotfill \pageref{sec:violin-absolute-color}
        \item \ref{sec:evaluate-violin-absolute}: Evaluation on Violin-Absolute-Color \dotfill \pageref{sec:evaluate-violin-absolute}
        \item \ref{sec:finetune-violin-absolute}: Fine-tuning on Zero-Entropy Color Variations \dotfill \pageref{sec:finetune-violin-absolute}
        \begin{itemize}
            \item \ref{sec:finetune-violin-absolute-detail}: Details \dotfill \pageref{sec:finetune-violin-absolute-detail}
            \item \ref{sec:finetune-violin-absolute-results}: Fine-tuning Results \dotfill \pageref{sec:finetune-violin-absolute-results}
            \item \ref{sec:finetune-t2i}: Performance on General T2I \dotfill \pageref{sec:finetune-t2i}
        \end{itemize}
        \item \ref{sec:violin-absolute-generalize}: Generalization Evaluation \dotfill \pageref{sec:violin-absolute-generalize}
    \end{itemize}
    \item \textbf{Section \ref{sec:discussion}:Discussions} \dotfill \pageref{sec:discussion}
    \item \textbf{Section \ref{sec:limitations}:Limitations} \dotfill \pageref{sec:limitations}
    \item \textbf{Section \ref{sec:social-impact}:Social Impact: AI Obedience and Reliability} \dotfill \pageref{sec:social-impact}

\end{itemize}

\newpage

\onecolumn

\section{Related Work}
\label{sec:relation}
\paragraph{Generative Models in Visual Domain}

Unlike the open-ended nature of text, the pixel-level determinism in visual generation makes it an ideal testbed for evaluating fine-grained instruction obedience. Capitalizing on this potential, visual generation, particularly image generation, has witnessed remarkable progress, with models now capable of synthesizing highly complex scenes that demand precise control. In text-to-image (T2I) generation, diffusion models\cite{ho2020denoising, ramesh2022hierarchical} have surpassed conventional generative approaches such as GAN\cite{goodfellow2020generative} and VAE\cite{kingma2013auto} by learning Gaussian denoising through reverse diffusion processes. Models such as Stable Diffusion\cite{rombach2022high, esser2024scaling}, FLUX\cite{labs2025flux}, and Qwen-Image\cite{wu2025qwen} have demonstrated exceptional performance with expanding training data and model scales.
Besides, unified models integrating multimodal understanding and visual generation have also emerged as a prominent research direction. Several studies adopt autoregressive (AR) frameworks analogous to LLMs\cite{team2024chameleon}, while others investigate hybrid architectures combining diffusion and AR\cite{xie2024show} to leverage complementary strengths of both paradigms. These models have achieved great breakthroughs in vision-language alignment and multi-modal generation tasks.

\paragraph{Low-Level AI Obedience} Existing research focuses on low-level visual obedience at the Semantic (Level-1) and Relational (Level-2) tiers. To achieve Level-1, foundational mechanisms like Classifier-Free Guidance (CFG) \cite{ho2022classifier} and ControlNet \cite{zhang2023adding} enforce alignment via latent guidance and spatial-semantic priors, while ImageReward \cite{xu2023imagereward} utilizes human preference learning to refine semantic fidelity. Level-2 research further improves attribute-object binding, using precise color codes \cite{butt2024colorpeel}, reference images~\cite{shum2025color}, or specialized modules to associate textures and quantities without attribute leakage \cite{li2024controlar, binyamin2025make}. However, these methods prioritize visual ``naturalness'' over perfect ``accuracy'', lacking the strict pixel-level control required for deterministic, zero-variance alignment with user specifications. ImagenWorld~\cite{sani2026imagenworld} tests instructions following in real-world scenes, exploring the breadth of complex semantics covering L1~L2. \cite{ramachandran2025well} shows that models like GPT-4o perform worse on geometric tasks than on semantic ones, often causing spatial hallucinations. This supports our argument that "architectural bias" makes models prefer rich semantics over precise pixel-level constraints (L4).

\paragraph{Obedience Benchmarks}
Multiple benchmarks have been proposed to evaluate the obedience of generative models across various dimensions. In terms of semantic and relational understanding, comprehensive suites like T2I-CompBench~\cite{huang2023t2i}, and GenAI-Bench~\cite{jiang2024genai} assess models' abilities in multi-object association, counting, and attribute binding. Regarding color specifically, some works are designed to evaluate the generation or reasoning ability of color in natural scenes~\cite{liang2025colorbench}. However, these benchmarks primarily assess low-level obedience, where visual plausibility often lacks true precision. They fail to evaluate the model's ability to prioritize strict instructions over its internal generative priors, which remains the focus of our study.

While recent benchmarks like VLRewardBench~\cite{li2025vl} and GenAI Arena~\cite{jiang2024genai} evaluate broad instruction-following and alignment, they primarily focus on semantic or aesthetic preferences. Violin complements these by targeting the Level-4 Obedience limit—deterministic, zero-entropy pixel control—which remains a critical failure mode not captured by existing reward-based or arena-style evaluations.

\paragraph{Harness Engineering}
Harness Engineering is a paradigm shift that defines an AI Agent as the synergy between a foundation model and its "harness", the structured scaffolding of memory, tools, and sandboxed environments that enforce reliability. The field was catalyzed by the seminal OpenAI report, which established that engineering the environment is as critical as training the model for complex tasks. Recent progress is further defined by~\cite{202603.1756}, which introduces a standardized architecture for agentic systems. Additional advancements, such as Meta-Harness~\cite{lee2026meta}, demonstrate how these scaffolds can now be automatically tuned, shifting the focus of AI development from prompt-tuning to the systematic engineering of deterministic infrastructures.

While Harness Engineering primarily explores various facets of external control and system-level optimization to refine model performance, the concept of Obedience we propose represents a more fundamental control capability that directly reflects the model's intrinsic proficiency acquired through training. We posit that Obedience serves as the bedrock of Harness Engineering and is a prerequisite for implementing more effective harnesses; ultimately, external scaffolding can only achieve its full potential when the underlying model possesses high inherent controllability.

\begin{table}[t]
\centering
\caption{Representative prompt examples for each task in Violin}
\label{tab:prompt_examples}
\begin{tabular}{>{\centering\arraybackslash}m{0.2\textwidth} >{\centering\arraybackslash}p{0.7\textwidth}}
\toprule
\textbf{Task} & \textbf{Example Prompts} \\
\midrule

Color Purity Var-1 & 
\begin{minipage}[c]{0.7\textwidth}
    \vspace{6pt}
    \begin{itemize}[leftmargin=*, itemsep=0.5\baselineskip, parsep=0pt, topsep=0pt, partopsep=0pt]
        \item "Generate an image with pure pink color."
        \item "Create a featureless entirely filled with red color."
        \item "Generate a perfectly uniform image in pink color. No texture, no gradient, no shadows."
    \end{itemize}
    \vspace{6pt}
\end{minipage} \\ 
\cdashline{1-2}

Color Purity Var-2 & 
\begin{minipage}[c]{0.7\textwidth}
    \vspace{6pt}
    \begin{itemize}[leftmargin=*, itemsep=0.5\baselineskip, parsep=0pt, topsep=0pt, partopsep=0pt]
        \item "Create an image divided vertically into two even halves: left in reddish purple, right in yellowish pink."
        \item "Produce a background with the top half in yellowish brown and the bottom half in greenish blue. Colors must remain flat and consistent throughout."
        \item "Create a flat background divided horizontally: lower half in violet color, upper half in black color. Each section should remain perfectly solid."
    \end{itemize}
    \vspace{6pt}
\end{minipage} \\ 
\cdashline{1-2}

Image-Mask & 
\begin{minipage}[c]{0.7\textwidth}
    \vspace{6pt}
    \begin{itemize}[leftmargin=*, itemsep=0.5\baselineskip, parsep=0pt, topsep=0pt, partopsep=0pt]
        \item "Apply the binary mask <mask> to the image <image>. For every pixel, if the mask value is 1, keep the original color from the image; if the mask value is 0, change it to pure black."
        \item "Perform a pixel-wise multiplication between image <image> and mask <mask>. Use the mask as a filter where 1 represents areas to be preserved and 0 represents areas to be changed to black. Output the resulting image."
        \item "Create an output image by mapping pixels from <image> onto binary mask <mask>: if a mask pixel is 1, use the source image color; if a mask pixel is 0, use pure black."
    \end{itemize}
    \vspace{6pt}
\end{minipage} \\ 
\cdashline{1-2}

Geometric Task & 
\begin{minipage}[c]{0.7\textwidth}
    \vspace{6pt}
    \begin{itemize}[leftmargin=*, itemsep=0.5\baselineskip, parsep=0pt, topsep=0pt, partopsep=0pt]
        \item "Generate a precise image of a solid white circle on a black background. Its diameter must be equal to 20\% of the image width, with its bounding box aligned with the left edge, vertically centered."
        \item "Create an image featuring a single solid white axis-aligned square on a black background, with a side length of 20\% image width, with its bounding box aligned with the left edge, vertically centered."
        \item "Generate a precise image of a solid white upright equilateral triangle on a black background. Its side length must be equal to 20\% of the image width, with its bounding box aligned with the left edge, vertically centered. The edges must be perfectly sharp with no textures."
    \end{itemize}
    \vspace{6pt}
\end{minipage} \\
\bottomrule
\end{tabular}
\end{table}

\section{Obedience Levels of Existing Literatures} 
\label{sec:category}
We categorize current benchmarks and metrics as follows:

\begin{itemize}[topsep=2pt, parsep=0pt]
    \item \textbf{Level-1}: 
    Recent benchmarks include DrawBench~\cite{saharia2022photorealistic} and PAINTSKILLS~\cite{cho2023dall}, alongside foundational alignment metrics such as CLIPScore~\cite{radford2021learning} BLIPScore~\cite{li2022blip}, Semantic Object Accuracy~\cite{hinz2020semantic} and VIdeoCon\cite{bansal2024videocon}.
    
    \item \textbf{Level-2}:
    Benchmarks like VISOR~\cite{gokhale2022benchmarking}, T2I-CompBench~\cite{huang2023t2i}, TIFA~\cite{hu2023tifa}, and SeeTrue~\cite{yarom2023you}, alongside metrics including SR-Score~\cite{huang2023t2i} for spatial logic, TIAM~\cite{grimal2024tiam} for attribute binding, and TIFA-Count~\cite{hu2023tifa} for numerical accuracy.

    \item \textbf{Level-3}
     A representative benchmark is \textbf{NEG-TTOI} \cite{cai2025tng}, which evaluates ``Not-to-Generate'' scenarios. 
    
\end{itemize}

\section{Other Obedience Example Tasks}
\label{sec:other-tasks}
To evaluate the deterministic control of generative models more comprehensively, we extend our experiments to three additional tasks: Area Ratio Control, Pixel-level Border Alignment, and Discrete Point Counting. These tasks are designed to test the model's ability to follow precise mathematical constraints rather than artistic interpretations.

To better highlight the persistence of these issues in state-of-the-art systems, we prioritize reporting results from Seedream-5.0, which currently stands as one of the most powerful and representative generative models. Despite its advanced capabilities, it consistently struggles with the deterministic constraints of our tasks. We also performed preliminary tests on GPT-Image-2, where we observed relatively fewer obedience issues. However, the systemic failures in Seedream-5.0 more clearly demonstrate that even top-tier models remain vulnerable to the "Paradox of Simplicity," prioritizing learned "aesthetic patterns" over precise user commands. 

(1) Area Ratio Control: We instruct the model to generate a canvas where exactly 13\% of the pixels on the left are pure blue (\#0000FF), and the remaining 87\% on the right are pure red (\#FF0000). Most models fail by creating a 50/50 symmetrical split or adding forbidden gradients at the boundary.

(2) Pixel-level Border Alignment: The model is required to produce a 1024x1024 black image with a 20-pixel-wide white border at the edge. Models typically fail this task by generating borders with inconsistent thickness or adding "aesthetic" glowing effects that violate the pixel-level requirement.

(3) Discrete Point Counting: We task the model with generating a black background containing exactly 10 white dots, each being a 4x4 pixel square. Due to probabilistic sampling noise, models often "over-generate," resulting in a random number of dots (e.g., 15 or more) instead of the requested amount.

\begin{figure}[t]
    \centering
    \includegraphics[width=0.88\linewidth]{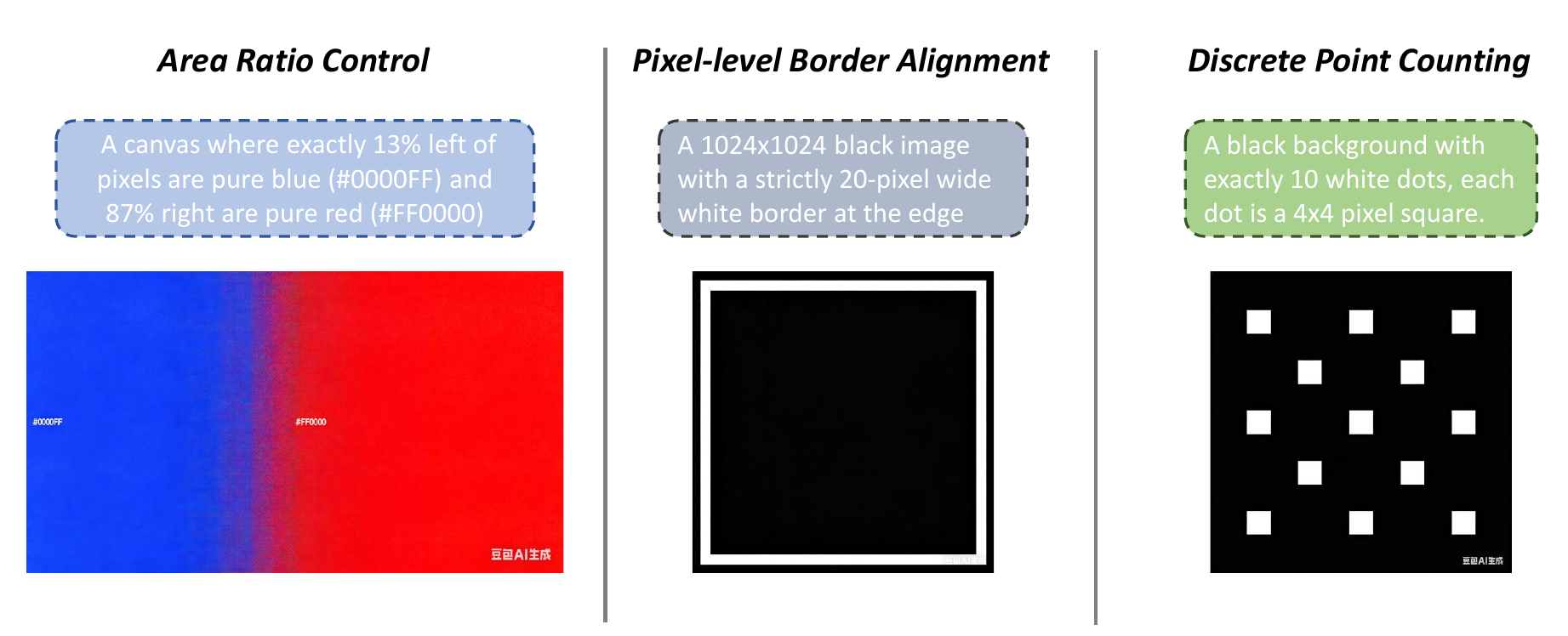}
    \caption{Failure cases on other low-entropy deterministic tasks.}
    \label{fig:additional_tasks}
\end{figure}

As shown in Fig.~\ref{fig:additional_tasks}, we provide several failure cases that highlight the gap between user instructions and model performance. These examples demonstrate that even advanced models tend to prioritize generating symmetrical and aesthetically pleasing images over following precise mathematical constraints. This "aesthetic inertia" causes the models to ignore the user's specific requirements in favor of learned visual patterns, which are consistent with our points.

\section{Benchmark Details.}
\label{sec:benchmark}
\subsection{Datasets}
\label{sec:dataset}
The evaluation dataset contains 7,745 samples in total, all generated at a standardized resolution of 512 $\times$ 512 pixels. Color Purity Variation 1 Task(Single Block) consists of 435 samples derived from 15 prompts, while Color Purity Variation 2 (Dual Blocks) includes 1,160 samples from 20 prompts. The Geometric Shape Generation Task utilizes 10 prompts to produce 750 samples focusing on \textbf{spatial positioning} and geometric fidelity. The Image Masking Task is the largest component, comprising 5,400 samples. This is constructed by selecting 1,800 base images from the BrushNet dataset~\cite{ju2024brushnet}, each evaluated against three mask types (Inpainting, Outpainting, and Random), resulting in 1,800 samples per mask configuration. Examples are shown in 
Tab.~\ref{tab:prompt_examples}. All images are programmatically synthesized at a fixed resolution of $512 \times 512$ pixels, with every pixel strictly adhering to the requested specification.

\begin{table*}[t]
\centering
\caption{Color Names, RGB, and HEX Values in ISCC-NBS system Level 2}
\label{tab:color_names}
\begin{tabular}{lll|lll}
\toprule
\textbf{Color Name} & \textbf{RGB} & \textbf{HEX} & \textbf{Color Name} & \textbf{RGB} & \textbf{HEX} \\
\midrule
Pink & (230, 134, 151) & \#E68697 & Red & (185, 40, 66) & \#B92842 \\
Yellowish pink & (234, 154, 144) & \#EA9A90 & Reddish orange & (215, 71, 42) & \#D7472A \\
Reddish brown & (122, 44, 38) & \#7A2C26 & Orange & (220, 125, 52) & \#DC7D34 \\
Brown & (127, 72, 41) & \#7F4829 & Orange yellow & (227, 160, 69) & \#E3A045 \\
Yellowish brown & (151, 107, 57) & \#976B39 & Yellow & (217, 180, 81) & \#D9B451 \\
Olive brown & (127, 97, 41) & \#7F6129 & Greenish yellow & (208, 196, 69) & \#D0C445 \\
Olive & (114, 103, 44) & \#72672C & Yellow green & (160, 194, 69) & \#A0C245 \\
Olive green & (62, 80, 31) & \#3E501F & Yellowish green & (74, 195, 77) & \#4AC34D \\
Green & (79, 191, 154) & \#4FBF9A & Bluish green & (67, 189, 184) & \#43BDB8 \\
Greenish blue & (62, 166, 198) & \#3EA6C6 & Blue & (59, 116, 192) & \#3B74C0 \\
Purplish blue & (79, 71, 198) & \#4F47C6 & Violet & (120, 66, 197) & \#7842C5 \\
Purple & (172, 74, 195) & \#AC4AC3 & Reddish purple & (187, 48, 164) & \#BB30A4 \\
Purplish pink & (229, 137, 191) & \#E589BF & Purplish red & (186, 43, 119) & \#BA2B77 \\
White & (231, 225, 233) & \#E7E1E9 & Gray & (147, 142, 147) & \#938E93 \\
Black & (43, 41, 43) & \#2B292B & & & \\
\bottomrule
\end{tabular}
\end{table*}

For the color purity task (var-1, var-2), we selected colors from the ISCC-NBS system, which defines three levels of color granularity. In this study, we use Level-2 colors, which consist of 29 common color categories (see Tab.~\ref{tab:color_names}). We then developed prompt templates with increasingly strict requirements.

For the image masking task, we designed three prompt templates, as shown in Tab.~\ref{tab:prompt_examples}. These templates are universal and contain no placeholders, meaning they are distinct and can be applied directly to all images. We selected natural images and their corresponding masks from the BrushNet dataset, as it provides high-quality semantic masks for our task.

For the geometric shape generation task, we selected fundamental shapes, specifically squares, circles, and equilateral triangles, that can be defined by a single parameter (e.g., side length or diameter). To ensure scale invariance and avoid potential data artifacts associated with absolute pixel values, we defined dimensions as percentages of the image. As illustrated in Tab.~\ref{tab:prompt_examples}, the task requires generating a centered shape of a fixed size aligned with the boundaries. This task is inherently zero-entropy, as it possesses a single, mathematically unique ground truth.

\subsection{Metrics}
\label{sec:metric}
\textbf{Notation.} Let $I \in \mathbb{R}^{W \times H \times 3}$ denote an image with width $W$ and height $H$, and $N = W \times H$ be the total number of pixels. For a pixel index $i \in \{1, \dots, N\}$ and color channel $c \in \{R, G, B\}$, $I_i^c$ represents the intensity value, and $Y(I_i)$ denotes the grayscale luminance. We define $M \in \{0, 1\}^{W \times H}$ as a binary mask. The operator $\ominus$ denotes morphological erosion, and $\mathcal{B}(M, d) = M \ominus (M \ominus \mathcal{K}_d)$ represents a boundary zone of width $d$. $F_{\text{shift}}(u, v)$ and $G(I)$ denote the centered 2D FFT spectrum and gradient magnitude, respectively.  In addition to previous definitions, let $\mathcal{C}$ denote the set of contour points extracted via the Suzuki algorithm. For a contour $\mathcal{C}$, we define its area as $A(\mathcal{C})$ and its image-relative centroid as $\mathbf{m} = (\bar{x}/W, \bar{y}/H)$, where $(\bar{x}, \bar{y})$ are the spatial moments. $\psi(\mathcal{C})$ represents the Hu invariant moments used for scale-invariant shape matching.

\paragraph{Color Purity Metric}
To quantitatively evaluate the color purity of the generated results, we employ five complementary metrics. Here, we detail the specific calculation formulations and implementation of the evaluation metrics.

(1). Euclidean Distance ($d_{\text{rgb\_ed}}$): This metric measures the absolute geometric distance between two color points in the linearized RGB color space:
\begin{equation}
d_{\text{rgb\_ed}} = \sqrt{\Delta R^2 + \Delta G^2 + \Delta
B^2}
\end{equation}
where $\Delta R, \Delta G,$ and $\Delta B$ signify the intensity differences between the target and reference samples across the three primary color channels.

(2). CIEDE2000 ($d_{\text{lab\_00}}$): this metric addresses the non-uniformity of the RGB space. By transforming coordinates into the CIELAB space and applying complex non-linear weighting functions, it accurately reflects the perceptual distance perceived by the human visual system. The detailed formulation can be seen in \cite{sharma2005ciede2000}.

(3). Standard Deviation ($d_{\text{sd}}$): This metric characterizes the spatial uniformity and color consistency within the image, formulated as:
\begin{equation}
\mu_c = \frac{1}{N} \sum_{i=1}^{N} I_i^c, \quad d_{\text{sd}} =
\frac{1}{3} \sum_{c \in \{R,G,B\}} \sqrt{\frac{1}{N} \sum_{i=1}^{N}
(I_i^c - \mu_c)^2}
\end{equation}

(4) Canny Edge Density ($d_{\text{ced}}$): To evaluate structural complexity and potential artifacts, we calculate the proportion of edge pixels:
\begin{equation}
d_{\text{ced}} = \frac{1}{N} \sum_{x=0}^{W-1} \sum_{y=0}^{H-1}
\mathbb{I}(E_{\text{canny}}(x, y) > 0)
\end{equation}
where $E_{\text{canny}}(x, y)$ is the binary edge map generated via the Canny operator, and $\mathbb{I}(\cdot)$ denotes the indicator function. This metric is sensitive to intensity variations, suitable for assessing the structural textures or shapes.

(5). High-Frequency Component Ratio ($d_{\text{hf}}$): By transforming the spatial image into the frequency domain via a 2D Fast Fourier Transform (FFT), we quantify the energy distribution. This ratio measures the dominance of high-frequency details relative to the total spectral energy:
\begin{equation}
d_{\text{hf}} = \frac{\sum \sum_{(u,v) \in
\Omega_{\text{high}}} |F_{\text{shift}}(u, v)|^2}{\sum_{u=0}^{W-1}
\sum_{v=0}^{H-1} |F_{\text{shift}}(u, v)|^2}
\end{equation}
where $F_{\text{shift}}(u, v)$ represents the zero-frequency centered 2D FFT spectrum, and $\Omega_{\text{high}}$ denotes the predefined high-frequency spectral domain. This corresponds to fine details and rapid variations, enabling the capture of subtle purity defects such as minor noise and local color discrepancies.

Then, we compute the mean as a unified metric:
\begin{equation}
\begin{aligned}
    d_{color\_mean} = (\hat d_{rgb\_ed} +\hat d_{lab\_00} + \hat d_{sd} +\hat d_{ced} +\hat d_{hf})/5\\
\end{aligned}
\end{equation}
where hat indicates that the variable has been normalized.

\paragraph{Image Mask Metric}
These metrics evaluate how strictly the synthesized image adheres to the input mask $M_{gt}$:

(1). Boundary IoU ($d_{\text{biou}}$): Focuses on contour alignment by computing the Intersection over Union within the boundary band $\mathcal{B}(M, d)$:
\begin{equation}
d_{\text{biou}} = 1.0 - \frac{|\mathcal{B}(M_{gen}, d) \cap \mathcal{B}(M_{gt}, d)|}{|\mathcal{B}(M_{gen}, d) \cup \mathcal{B}(M_{gt}, d)|}
\end{equation}

(2). Content Leakage ($d_{\text{leak}}$): Evaluates the undesired "seepage" of luminance into the masked regions. It is defined as the normalized mean intensity within the eroded mask core $\hat{M}_{gt}$:
\begin{equation}
    d_{\text{leak}} = \frac{1}{|\hat{M}_{gt}|} \sum_{i \in \hat{M}_{gt}} \frac{Y(I_{gen, i})}{255}
\end{equation}

(3). Edge Transition Consistency ($d_{\text{edge}}$): Measures the preservation of sharpness at the mask interface by comparing the relative gradient magnitude within the edge zone $\mathcal{E}$:
\begin{equation}
    d_{\text{edge}} = \max\left(0, 1 - \frac{\sum_{j \in \mathcal{E}} G(I_{gen, j})}{\sum_{j \in \mathcal{E}} G(I_{gt, j})}\right)
\end{equation}

(4). Shape Fidelity ($d_{\text{iou}}$ \& $d_{\text{dist}}$): We further employ standard IoU and distance transform-based metrics to ensure global area coverage and penalize spatial deformations between the synthesized shape and the reference mask.

Similarly, we compute the mean as a unified metric:
\begin{equation}
\begin{aligned}
    d_{shape\_mean} = (\hat d_{iou} +\hat d_{dist} + \hat d_{size} +\hat d_{shape} +\hat d_{pure})/5\\
\end{aligned}
\end{equation}

\paragraph{Geometric Shape Metric}
To evaluate the model's ability to synthesize specific geometries at designated locations and scales, we employ the following normalized metrics:

(1). Normalized Centroid Distance ($d_{\text{dist}}$): Measures the spatial displacement between the synthesized object and the ground truth in a normalized coordinate space:
\begin{equation}
d_{\text{dist}} = \min\left(\frac{|\mathbf{m}{gen} - \mathbf{m}{gt}|_2}{\sqrt{2}}, 1.0\right)
\end{equation}

(2). Normalized Size Deviation ($d_{\text{size}}$): Evaluates the relative area ratio between the generated shape and the reference, ensuring scale consistency:
\begin{equation}
    d_{\text{size}} = \min\left(\left| \frac{A(\mathcal{C}_{gen}) / N_{gen}}{A(\mathcal{C}_{gt}) / N_{gt}} - 1.0 \right|, 1.0\right)
\end{equation}
where $N$ is the total pixel count of the respective canvas.

(3). Shape Complexity Matching ($d_{\text{shape}}$): Quantifies the topological similarity using Hu invariant moments. This metric is inherently invariant to translation, scale, and rotation, focusing purely on the geometric primitive's structure:
\begin{equation}
    d_{\text{shape}} = \min\left(\text{Match}(\psi_{gen}, \psi_{gt}), 1.0\right)
\end{equation}

(4). In-situ Color Purity ($d_{\text{purity}}$): Distinct from global purity, this metric calculates the average standard deviation of RGB channels specifically within the generated object's mask area $\mathcal{M}_{gen}$:
\begin{equation}
    d_{\text{purity}} = \min\left(\frac{1}{128} \cdot \text{std}_{i \in \mathcal{M}_{gen}}(I_{gen, i}), 1.0\right)
\end{equation}

(5). Rescaled Shape IoU ($d_{\text{iou}}$): To perform pixel-level alignment analysis, the generated mask is rescaled to the ground truth dimensions using nearest-neighbor interpolation before computing the Intersection over Union. This measures the final aligned overlap fidelity.

Similarly, we compute the mean as a unified metric:
\begin{equation}
\begin{aligned}
    d_{mask\_mean} = (\hat d_{iou} +\hat d_{dist} + \hat d_{biou} +\hat d_{leak} +\hat d_{edge})/5\\
\end{aligned}
\end{equation}

\paragraph{Multi-Block Metric}
For multi-color block cases, direct application of the above metrics would introduce evaluation bias, as the expected color boundaries and regional variations would be identified as defects. To mitigate this, we employ a block-wise evaluation strategy: the multi-color block image is first partitioned into uniform segments by color regions, metrics are then computed independently for each sub-region, and the results are averaged to yield the final image score.

\section{Supplementary Zero-Entropy Color Task}
\label{sec:ze-color}
Considering that color generation is the most fundamental task, it is noteworthy that color descriptions in natural language, as used in our main study, Violin, do not exhibit "zero entropy" but rather a low-entropy state due to linguistic variance. While Violin initially adopts natural language to account for the scarcity of precise color codes in large-scale training datasets, we introduce Violin-Absolute-Color as a critical extension to investigate scenarios of absolute zero entropy and extreme precision. In this benchmark, hexadecimal (Hex) codes are employed as color descriptors. By incorporating six distinct variations, we conducted extensive experiments that yielded several compelling insights into the limits of model performance under deterministic constraints. Examples can be seen in Tab.~\ref{tab:zero_entropy_prompt_examples}.

\subsection{Violin-Absolute-Color Benchmark}
\label{sec:violin-absolute-color}
\textbf{Variation 1 and Variation 2}
Following the Violin framework, these variations utilize single-block and double-block configurations. The primary distinction lies in our adoption of hexadecimal color codes for representation, which are systematically derived from the ISCC-NBS color system.

\textbf{Variation 3: Four-Color Block}
We extend the evaluation to four-color block tasks (arranged in a $2 \times 2$ grid) to further examine the models' capacity for handling complex color contrasts. This specific configuration is selected to provide unambiguous spatial partitioning, thereby avoiding the inherent layout ambiguity often present in other arrangements, such as three-color blocks. 

\textbf{Variation 4: Fuzzy Color}
Building on the variation-1 scenario, we provide an acceptable color range rather than exact values to evaluate model obedience under fuzzy constraints. This variation systematically assesses how models handle instruction-following precision when given flexible yet bounded color specifications.

\textbf{Variation 5: Multilingual Prompts}
To evaluate model obedience across diverse linguistic systems, we extend the variation-1 to Chinese and French, representing logographic and Latin-based writing systems, respectively. This expansion allows us to investigate whether differences in grammatical structure, color vocabulary, and cultural context impact the model's precision in executing color-related instructions.

\textbf{Variation 6: Multiple Color Space}
To assess the impact of color encoding, we extend the single-color task to include RGB and HSL formats alongside hex codes. This expansion allows us to investigate whether varying mathematical foundations across these representation methods introduce specific biases or affect the model's depth of understanding.

\begin{table}[t]
\centering
\caption{Representative prompt examples for each variation in  Violin-Absolute-Color Benchmark.}
\label{tab:zero_entropy_prompt_examples}
\begin{tabular}{p{2cm}p{11cm}}
\toprule
Variation & Example Prompt \\
\midrule
1 & Generate an image with pure color \#D9B451 (Hex code). \\
2 & Specified in Hex code, produce a background divided horizontally into two equal halves: top color \#72672C and bottom color \#D9B451. \\
3 & Specified in Hex code, generate an image divided into four quadrants: top-left color \#7F4829, top-right color \#B92842, bottom-left color \#7F4829, bottom-right color \#3B74C0. No gradients or shadows. \\
4 & Specified in Hex code, generate an image in a single solid color within the range of \#58321D and \#A65E35. \\
5 & \begin{CJK}{UTF8}{gbsn}
生成一张颜色为 \#B92842（十六进制代码）的纯色图像。
\end{CJK}
 \\
6 & Generate a plain background using color rgb(217, 180, 81). \\
\bottomrule
\end{tabular}
\end{table}

\subsection{Evaluation on Violin-Absolute-Color}
\label{sec:evaluate-violin-absolute}
\begin{table}[t]
    \centering
    \small
    \setlength{\tabcolsep}{8pt}

    \caption{Comprehensive Evaluation on Violin-Absolute-Color (Var 1-6)}
    \begin{tabular}{llccccc >{\columncolor{gray!15}}c}
        \toprule
        Var & Models & {rgb-ed} & {lab-00} & {sd} & {ced} & {hf} & color-mean \\
        \midrule
        \multirow{5}{*}{Var-1} 
        & SANA          & 0.288 & 0.331 & 0.232 & 0.028 & 0.030 & 0.182 \\
        & Janus-Pro-1.5 & 0.344 & 0.410 & 0.193 & 0.006 & 0.004 & 0.191 \\
        & FLUX.1        & 0.364 & 0.387 & 0.044 & 0.001 & 0.001 & 0.159 \\
        & Qwen-Image    & 0.156 & 0.180 & 0.058 & 0.002 & 0.021 & 0.083 \\
        & OmniGen2      & 0.397 & 0.402 & 0.202 & 0.070 & 0.016 & 0.217 \\
        \midrule 
        \multirow{5}{*}{Var-2}
        & SANA          & 0.315 & 0.352 & 0.312 & 0.019 & 0.021 & 0.204 \\
        & Janus-Pro-1.5 & 0.335 & 0.359 & 0.216 & 0.009 & 0.012 & 0.186 \\
        & FLUX.1        & 0.357 & 0.384 & 0.117 & 0.005 & 0.005 & 0.174 \\
        & Qwen-Image    & 0.128 & 0.143 & 0.111 & 0.007 & 0.028 & 0.083 \\
        & OmniGen2      & 0.388 & 0.396 & 0.236 & 0.036 & 0.026 & 0.216 \\
        \midrule 
        \multirow{5}{*}{Var-3}
        & SANA          & 0.369 & 0.401 & 0.433 & 0.030 & 0.035 & 0.254 \\
        & Janus-Pro-1.5 & 0.338 & 0.383 & 0.220 & 0.022 & 0.024 & 0.197 \\
        & FLUX.1        & 0.386 & 0.416 & 0.142 & 0.008 & 0.008 & 0.192 \\
        & Qwen-Image    & 0.229 & 0.264 & 0.174 & 0.015 & 0.029 & 0.142 \\
        & OmniGen2      & 0.410 & 0.433 & 0.242 & 0.024 & 0.025 & 0.227 \\
        \midrule
        \multirow{5}{*}{Var-4} 
        & SANA          & 0.321 & 0.367 & 0.202 & 0.048 & 0.017 & 0.191 \\
        & Janus-Pro-1.5 & 0.281 & 0.380 & 0.086 & 0.005 & 0.004 & 0.151 \\
        & FLUX.1        & 0.336 & 0.382 & 0.062 & 0.001 & 0.001 & 0.156 \\
        & Qwen-Image    & 0.280 & 0.303 & 0.123 & 0.004 & 0.018 & 0.146 \\
        & OmniGen2      & 0.384 & 0.400 & 0.132 & 0.042 & 0.017 & 0.195 \\
        \midrule 
        \multirow{5}{*}{Var-5(C)}
        & SANA          & 0.304 & 0.344 & 0.178 & 0.041 & 0.011 & 0.176 \\
        & Janus-Pro-1.5 & 0.340 & 0.373 & 0.468 & 0.025 & 0.013 & 0.244 \\
        & FLUX.1        & 0.415 & 0.397 & 0.529 & 0.105 & 0.051 & 0.300 \\
        & Qwen-Image    & 0.145 & 0.178 & 0.081 & 0.005 & 0.013 & 0.084 \\
        & OmniGen2      & 0.397 & 0.414 & 0.173 & 0.051 & 0.015 & 0.210 \\
        \midrule
        \multirow{5}{*}{Var-5(F)}
        & SANA          & 0.306 & 0.342 & 0.257 & 0.031 & 0.041 & 0.195 \\
        & Janus-Pro-1.5 & 0.343 & 0.382 & 0.389 & 0.014 & 0.015 & 0.229 \\
        & FLUX.1        & 0.372 & 0.390 & 0.267 & 0.020 & 0.016 & 0.213 \\
        & Qwen-Image    & 0.161 & 0.201 & 0.051 & 0.002 & 0.007 & 0.084 \\
        & OmniGen2      & 0.428 & 0.431 & 0.379 & 0.116 & 0.024 & 0.276 \\
        \midrule 
        \multirow{5}{*}{Var-6(rgb)}
        & SANA          & 0.331 & 0.383 & 0.381 & 0.061 & 0.034 & 0.238 \\
        & Janus-Pro-1.5 & 0.343 & 0.406 & 0.250 & 0.003 & 0.005 & 0.201 \\
        & FLUX.1        & 0.359 & 0.394 & 0.153 & 0.001 & 0.002 & 0.182 \\
        & Qwen-Image    & 0.212 & 0.221 & 0.063 & 0.001 & 0.009 & 0.101 \\
        & OmniGen2      & 0.427 & 0.447 & 0.056 & 0.012 & 0.001 & 0.189 \\
        \midrule
        \multirow{5}{*}{Var-6(hsl)}
        & SANA          & 0.339 & 0.387 & 0.335 & 0.052 & 0.027 & 0.228 \\
        & Janus-Pro-1.5 & 0.355 & 0.396 & 0.264 & 0.007 & 0.006 & 0.206 \\
        & FLUX.1        & 0.366 & 0.388 & 0.139 & 0.003 & 0.002 & 0.180 \\
        & Qwen-Image    & 0.314 & 0.346 & 0.102 & 0.004 & 0.007 & 0.155 \\
        & OmniGen2      & 0.374 & 0.394 & 0.103 & 0.034 & 0.003 & 0.182 \\
        \bottomrule
    \end{tabular}
    \label{tab-ze-evaluate}
\end{table}

To systematically evaluate performance across different variations, we conducted a series of experiments on five representative open-source models using a consistent set of metrics. These specific models were selected for their documented proficiency in color-related tasks, thereby mitigating potential biases or artifacts stemming from suboptimal data processing. 

As in Tab.~\ref{tab-ze-evaluate}, Qwen-Image emerges as the top performer, consistently achieving the lowest values in rgb-ed and lab-00 across the majority of variations. For instance, it reaches minimums of 0.128 and 0.143 in Var-2, respectively, indicating superior precision in color prediction compared to the other evaluated models. In contrast, OmniGen2 yields significantly higher error metrics, reflecting a substantial deviation in handling absolute color tasks. Furthermore, model performance exhibits notable sensitivity to task formulations; for example, FLUX.1 performs exceptionally well in the sd metric under Var-1 (0.044) but sees a sharp degradation to 0.529 in Var-5(C). Overall, Qwen-Image demonstrates the most robust color alignment capabilities within this benchmark.

\subsection{Fine-tuning on Zero-Entropy Color Variations}
\label{sec:finetune-violin-absolute}
As similar attempts have been made in other benchmarks to verify the impact of specific data on task performance~\cite{huang2023t2i}. Accordingly, we conducted additional fine-tuning experiments, using the most fundamental and intuitive 'pure color' task, to validate data effectiveness. We must emphasize, however, that our benchmark does not intend for models to achieve high scores through specialized data tuning. Instead, it aims to reflect the emergence of intelligence during training and evaluate model capabilities under conventional data settings; this experiment serves solely as a supplement.

\subsubsection{Details}
\label{sec:finetune-violin-absolute-detail}
\paragraph{Setting} To assess learning capabilities on Violin-Absolute-Color522, we conduct fine-tuning experiments using an 8:2 random train-test split.
For open-source models, we apply LoRA for parameter-efficient fine-tuning over 3,000 steps to avoid overfitting. Closed-source models are evaluated only through direct inference via official APIs, as fine-tuning is unavailable.
For fine-tuning experiments, we primarily report results on Qwen-Image 
with the following configuration.

\textbf{LoRA Configuration.} We apply LoRA (Low-Rank Adaptation) for 
parameter-efficient fine-tuning with rank $r=16$, $\alpha=32$, and 
dropout rate of 0.1. LoRA modules are applied to the attention 
projection matrices (\texttt{to\_k}, \texttt{to\_q}, \texttt{to\_v}, 
and \texttt{to\_out.0}). The base model is quantized to qfloat8 to 
reduce memory consumption.

\textbf{Optimization.} We use the 8-bit AdamW optimizer with a learning 
rate of $3 \times 10^{-4}$ and a constant learning rate schedule 
without warmup. Gradient clipping is applied with a maximum norm of 1.0.

\textbf{Training Configuration.} Training is conducted on 4 GPUs with a 
per-device batch size of 8, resulting in an effective batch size of 32. 
Models are trained for 3,000 steps with bf16 mixed precision.

\textbf{Data Processing.} All training images are resized to $384 
\times 384$ resolution, due to the specific requirement of Janus-Pro. We precompute and cache text embeddings and image latents to accelerate training. A caption dropout rate of 0.1 is applied for regularization.

\subsubsection{Fine-tuning Results}
\label{sec:finetune-violin-absolute-results}
As shown in Tab.~\ref{tab-finetune-variations} and Tab.~\ref{tab-ze-finetune-models}, we conduct a detailed analysis of the impact of fine-tuning on model performance across different variations and architectures.

\paragraph{Performance across Variations.} Tab.~\ref{tab-finetune-variations} presents the fine-tuning results for the Qwen-Image model individually trained on six distinct variations (Var 1--6). Although fine-tuning was expected to adapt the model to specific tasks, the results indicate that the performance gains are remarkably limited. Across all variations, we observe a consistent decline in the error metrics (denoted in the "Change" column), yet the absolute performance remains at a relatively low scale level. For instance, while Var-3 and Var-4 show a more significant numerical decrease in error ($\downarrow 0.062$ and $\downarrow 0.035$ respectively), the overall color-mean does not reach a high-precision threshold. This suggests that while fine-tuning can offer marginal improvements on specific low-scale tasks, it struggles to fundamentally resolve the underlying challenges inherent in high-accuracy color perception.

\paragraph{Cross-Model Generalization.} In Tab.~\ref{tab-ze-finetune-models}, we investigate the effectiveness of fine-tuning on Variation 1 across several state-of-the-art open-source models, including SANA, Janus-Pro-1.5, FLUX.1, Qwen-Image, and OmniGen2. The empirical evidence reinforces our previous observation: fine-tuning yields varying but generally modest improvements. Janus-Pro-1.5 exhibits the most substantial error reduction ($\downarrow 0.146$), whereas models like Qwen-Image and OmniGen2 show almost negligible gains ($\downarrow 0.004$ and $\downarrow 0.005$). This disparity highlights that the "ceiling" for improvement via fine-tuning is heavily dependent on the base model's architecture. Overall, the fine-tuning approach appears insufficient to bridge the gap for more complex color-absolute tasks, providing only limited enhancements in narrow, small-scale scenarios.

\begin{table*}[t]
    \centering
    \small
    \setlength{\tabcolsep}{12pt} 
    \caption{\textbf{Fine-tuning analysis using Violin-Absolute-Color on the Qwen-Image model.} Results report the performance after fine-tuning the model individually on each variation (Var 1--6).}
    \begin{tabular}{lccccc >{\columncolor{gray!15}}c >{\columncolor{blue!10}}c} 
        \toprule
        \textbf{Models} & {rgb-ed} & {lab-00} & {sd} & {ced} & {hf} & \textbf{color-mean} & \textbf{Change}\\
        \midrule
        Var-1   & 0.093 & 0.124  & 0.023 & 0.001 & 0.001 & 0.048 & $\downarrow$ 0.035\\
        Var-2   & 0.088 & 0.110  & 0.027 & 0.003 & 0.005 & 0.046 & $\downarrow$ 0.037\\
        Var-3   & 0.163 & 0.186  & 0.042 & 0.006 & 0.005 & 0.080 & $\downarrow$ 0.062\\
        Var-4   & 0.237 & 0.266  & 0.052 & 0.001 & 0.003 & 0.111 & $\downarrow$ 0.035\\
        Var-5   & 0.103 & 0.140  & 0.023 & 0.001 & 0.004 & 0.054 & $\downarrow$ 0.030\\
        Var-6   & 0.190 & 0.227  & 0.011 & 0.001 & 0.001 & 0.086 & $\downarrow$ 0.042\\
        \bottomrule
    \end{tabular}
    \label{tab-finetune-variations}
\end{table*}

\begin{table*}[t]
    \centering
    \small
    \setlength{\tabcolsep}{10pt} 
    \caption{\textbf{Fine-tuning analysis using Violin-Absolute-Color on Variation 1 across different open-source models.} Results report the performance after fine-tuning the model individually.}
    \begin{tabular}{lccccc >{\columncolor{gray!15}}c >{\columncolor{blue!10}}c}
        \toprule
        \textbf{Models} & {rgb-ed} & {lab-00} & {sd} & {ced} & {hf} & \textbf{color-mean} & \textbf{Change} \\
        \midrule
        SANA           & 0.292 & 0.330 & 0.211 & 0.001 & 0.027 & 0.172 & $\downarrow$ 0.010 \\
        Janus-Pro-1.5  & 0.090 & 0.121 & 0.010 & 0.001 & 0.001 & 0.045 & $\downarrow$ 0.146 \\
        FLUX.1         & 0.227 & 0.283 & 0.004 & 0.000 & 0.001 & 0.103 & $\downarrow$ 0.056 \\
        Qwen-Image     & 0.167 & 0.208 & 0.018 & 0.001 & 0.001 & 0.079 & $\downarrow$ 0.004 \\
        OmniGen2       & 0.413 & 0.412 & 0.159 & 0.060 & 0.014 & 0.212 & $\downarrow$ 0.005 \\

        \bottomrule
    \end{tabular}
    \label{tab-ze-finetune-models}
\end{table*}

 \subsubsection{Performance on General T2I}
 \label{sec:finetune-t2i}
\begin{figure}
    \centering
    \includegraphics[width=0.74 \linewidth]{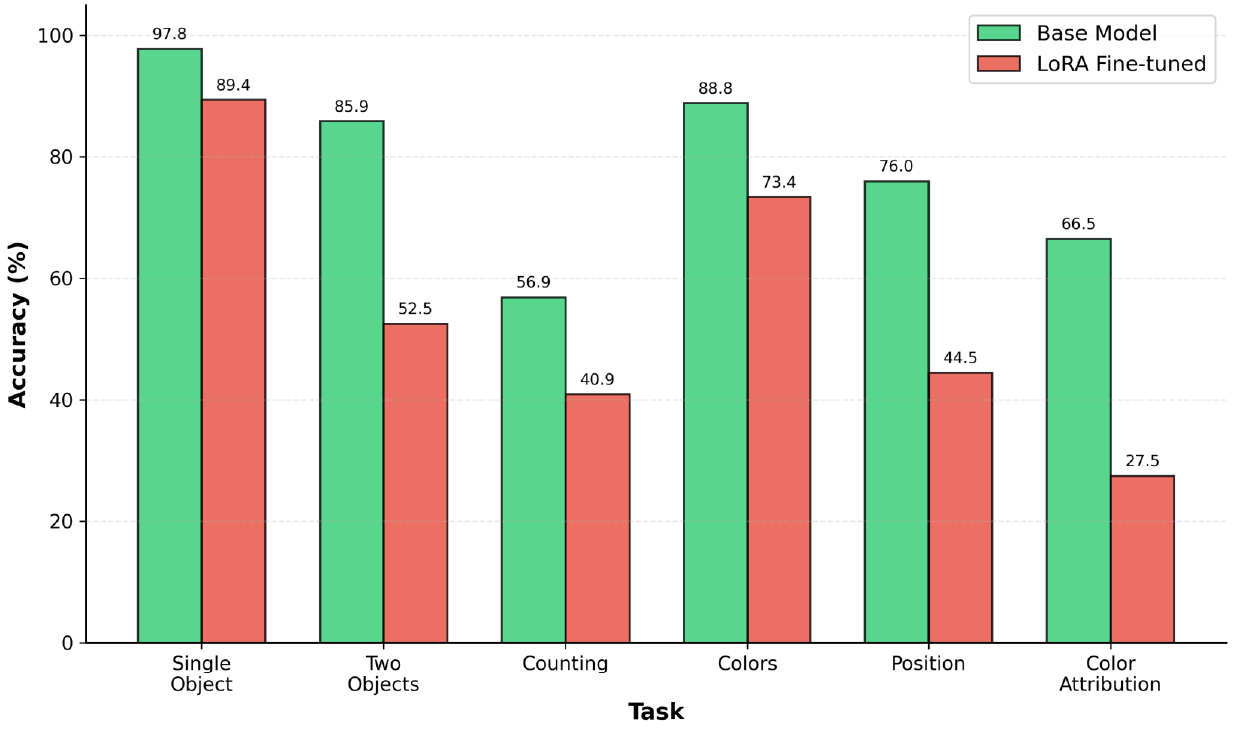}
    \caption{\textbf{Performance degradation on GenEval after fine-tuning.} Comparison between the base and LoRA fine-tuned models reveals a noticeable loss in general generative abilities across multiple semantic and spatial tasks.}
    \label{fig-dynamics}
\end{figure}

To investigate whether finetuning solely affects models' general generation capabilities, we track performance throughout the training process. Specifically, we evaluate models on GenEval~\cite{ghosh2023geneval} during fine-tuning, plotting performance as shown in Fig.~\ref{fig-dynamics}.

The figure provides critical evidence that simply augmenting data via fine-tuning may not be a viable solution for enhancing specific model capabilities. As illustrated, the LoRA fine-tuned model suffers a consistent and significant performance drop across all GenEval tasks—such as counting, positioning, and color attribution—compared to the base model.

This performance erosion suggests that even if our objective is to encourage the "emergence" of specific abilities through targeted data, the fine-tuning process tends to compromise the model's existing general synthesis capabilities. It further implies that simply adding data is not a panacea and may even be counterproductive in a fine-tuning context. Due to limited computational resources, we have not yet explored integrating such data during the pre-training phase from scratch. Whether fundamental integration at the pre-training level can mitigate this "general ability loss" remains a promising direction for future research.

\subsection{Generalization Evaluation}
\label{sec:violin-absolute-generalize}

Since the above experiments provide only a coarse-grained analysis, we conduct fine-grained evaluations of model generalization in color obedience. We design three data partition strategies: (1) prompt-based partitioning(Prompt-Split): splitting data by template to evaluate different instruction formulations; (2) localized hue partitioning(Hue-Split1): testing purple range ($280^\circ$-$320^\circ$ hue) while training on the remainder; (3) distributed hue partitioning(Hue-Split2)): training on three distributed ranges ($0^\circ$-$60^\circ$, $120^\circ$-$180^\circ$, $240^\circ$-$300^\circ$) with the rest for testing.

As shown in Tab.~\ref{tab-generalization-test}, while color purity generalizes across different settings, color precision exhibits bad performance. We hypothesize that current models may rely more on memorization rather than developing a truly robust understanding of the color space.

\begin{table*}[t]
    \centering
    \small
    \setlength{\tabcolsep}{12pt} 
    \caption{Evaluating the model's zero-shot generalization on three designed color partition strategies.}
    \begin{tabular}{lccccc >{\columncolor{gray!15}}c}
        \toprule
        \textbf{Models} & {rgb-ed} & {lab-00} & {sd} & {ced} & {hf} & \textbf{color-mean} \\
        \midrule
        Prompt-Split   & 0.095 & 0.113 & 0.048 & 0.004 & 0.010 & 0.054 \\
        Hue-Split1     & 0.127 & 0.156 & 0.023 & 0.003 & 0.002 & 0.062 \\
        Hue-Split2     & 0.128 & 0.158 & 0.034 & 0.003 & 0.004 & 0.065 \\
        \bottomrule
    \end{tabular}
    \label{tab-generalization-test}
\end{table*}

\section{Discussions}
\label{sec:discussion}
\textbf{The Atomic Nature of Pure Color Obedience.} We intentionally adopt pure color generation as the atomic unit for evaluating Level-4 obedience. In contrast to complex compositional tasks where failures are often conflated with layout ambiguities or attribute-binding errors, pure color provides an unambiguous, pixel-perfect metric. Our rationale is that if a model lacks the systemic precision required to control a single, deterministic RGB value, it inherently fails at the most fundamental limit of instruction following. By isolating this ``minimum viable obedience'', we provide a rigorous baseline for assessing deterministic control in generative AI.

\textbf{Inductive Bias and the ``Aesthetic Inertia''.} Our empirical findings suggest that the obedience gap is not merely a consequence of data scarcity, but is deeply rooted in the inductive bias of current generative architectures. Standard training objectives—such as maximizing likelihood or minimizing reconstruction loss—naturally incentivize models to converge toward high-entropy manifold regions characterized by rich textures and complex semantics. Generating a pure color image (a zero-entropy state) requires the model to actively suppress these learned priors. This phenomenon, which we term "Aesthetic Inertia," indicates that scaling data alone, without fundamental changes to architectural constraints or loss formulations, is unlikely to bridge the gap toward true Level-4 obedience.

\textbf{Benchmark Utility and Open-Sourcing}. To facilitate further exploration into the limits of AI controllability, we introduce the Violin benchmark as a diagnostic tool for the community. Violin is designed to be complementary to existing semantic-alignment benchmarks by focusing specifically on the deterministic precision of visual outputs. To support reproducibility and future research, we will release the complete Violin suite, including our automated evaluation pipeline and dataset.

\section{Limitations}
\label{sec:limitations}
While our study provides insights into model obedience, it has several limitations that suggest directions for future research:

(1) Model Scope: This research focuses primarily on mainstream, large-scale models. Because we did not test niche or highly specialized architectures, it is not yet clear if our findings apply to every type of model design.

(2) Mitigation Strategies: Our work concentrates on measuring and explaining obedience failures. We do not provide specific solutions, such as new training objectives or structural changes, to fix these issues. Finding ways to resolve these failures remains an open challenge.

(3) Sampling Parameters: All experiments were conducted using standard sampling settings (such as default temperature and top-p values). It is possible that different sampling configurations could change how obedience manifests, which should be explored in future studies.

 \section{Social Impact: AI Obedience and Reliability}
 \label{sec:social-impact}
This paper introduces the concept of AI Obedience and the Violin benchmark, which have significant social implications for the deployment of artificial intelligence:

Enhancing Safety in Critical Sectors: In high-stakes fields like medical imaging and autonomous driving, "deterministic control" is vital. Researching how to make models strictly follow instructions (e.g., precise pixel-level labeling) helps prevent system failures or misdiagnoses caused by AI hallucinations or random deviations.

Establishing Industry Trust: As AI moves into professional production workflows, predictability is the foundation of user trust. By quantifying how well models perform simple, objective tasks (like generating a solid color), this research provides a scientific basis for assessing AI reliability, shifting AI from "probabilistic expression" to "deterministic execution".

Mitigating Unintended Biases: The paper reveals that models often prioritize "aesthetic inertia" over precise technical constraints. Improving obedience ensures that AI serves as an objective tool rather than being swayed by hidden biases or stylistic preferences found in its training data, ensuring purer and more accurate outputs.

\end{document}